\documentclass{svmult}

\usepackage[caption=false]{subfig}
\usepackage{algorithm}      
\usepackage{algorithmicx}      
\usepackage[noend]{algpseudocode} 
\usepackage{graphicx}       
\usepackage{booktabs}       
\usepackage{amsmath}        
\usepackage{microtype}      
\usepackage[utf8]{inputenc} 
\usepackage{newtxmath}      
\usepackage{hyperref}       
\usepackage{acro}           
\usepackage[free-standing-units=true,round-mode=places,round-precision=1,detect-weight=true]{siunitx} 
\usepackage{svg}            
\usepackage{mathtools}      

\usepackage{url}
\usepackage[subtle]{savetrees}
\usepackage[sort]{cite}

\usepackage{makecell}      
\newcommand{\mat}[1]{\boldsymbol{#1}}     

\makeatletter
\renewcommand{\ALG@beginalgorithmic}{\small}
\makeatother

\algnewcommand{\LineComment}[1]{\State \(\triangleright\) #1}

\DeclareAcronym{AGP}{
  short = AGP,
  long  = art gallery problem
}
\DeclareAcronym{AGL}{
  short = AGL,
  long  = height above ground level
}
\DeclareAcronym{BCD}{
  short = BCD,
  long  = boustrophedon cell decompositon
}
\DeclareAcronym{CCD}{
  short = CCD,
  long  = optimal convex cell decomposition without Steiner points
}
\DeclareAcronym{CGAL}{
  short = CGAL,
  long  = the Computational Geometry Algorithms Library
}
\DeclareAcronym{DTM}{
  short = DTM,
  long  = digital terrain model
}
\DeclareAcronym{FOV}{
  short = FOV,
  long  = field of view,
  short-indefinite = an
}
\DeclareAcronym{GK}{
  short = GK,
  long  = Gutin and Karapetyan solver
}
\DeclareAcronym{GPS}{
  short = GPS,
  long  = Global Positioning Service
}
\DeclareAcronym{RTK}{
  short = RTK,
  long  = Real-Time Kinematic
}
\DeclareAcronym{E-GTSP}{
  short = E-GTSP,
  long  = Equality Generalized Traveling Salesman Problem,
  long-indefinite = an,
  short-indefinite = an
}
\DeclareAcronym{MAV}{
  short = MAV,
  long  = unmanned rotary-wing micro aerial vehicle,
  short-indefinite = an,
  long-indefinite = an
}
\DeclareAcronym{NFZ}{
  short = NFZ,
  long  = no-fly-zone,
  short-indefinite = an
}
\DeclareAcronym{PWH}{
  short = PWH,
  long  = general polygon with holes
}
\DeclareAcronym{ROS}{
  short = ROS,
  long  = robot operating system
}
\DeclareAcronym{SAR}{
  short = GPSAR,
  long  = ground-penetrating synthetic aperture radar
}
\DeclareAcronym{SHA}{
  short = SHA,
  long  = suspected hazardous area
}
\DeclareAcronym{TCD}{
  short = TCD,
  long  = trapezoidal cell decomposition
}
\DeclareAcronym{TSP}{
  short = TSP,
  long  = traveling salesman problem
}

\begin{document}
\title*{Revisiting Boustrophedon Coverage Path Planning as a Generalized Traveling Salesman Problem}
\author{
Rik B\"{a}hnemann, Nicholas Lawrance, Jen Jen Chung, Michael Pantic,\\ Roland Siegwart, and Juan Nieto}

\institute{Authors are with Autonomous Systems Lab, ETH Zurich, e-mail: \texttt{\small \{brik, lawrancn, chungj, mpantic, rsiegwart, nietoj\}@ethz.ch}}

\maketitle
\thispagestyle{empty}
\pagestyle{empty}
\abstract{
In this paper, we present a path planner for low-altitude terrain coverage in known environments with \acp{MAV}.
Airborne systems can assist humanitarian demining by surveying \acp{SHA} with cameras, \ac{SAR}, and metal detectors.
Most available coverage planner implementations for \acp{MAV} do not consider obstacles and thus cannot be deployed in obstructed environments.
We describe an open source framework to perform coverage planning in polygon flight corridors with obstacles.
Our planner extends boustrophedon coverage planning by optimizing over different sweep combinations to find the optimal sweep path, and considers obstacles during transition flights between cells.
We evaluate the path planner on \SI{320}{} synthetic maps and show that it is able to solve realistic planning instances fast enough to run in the field.
The planner achieves \SI{14}{\percent} lower path costs than a conventional coverage planner.
We validate the planner on a real platform where we show low-altitude coverage over a sloped terrain with trees.
}

\section{Introduction}
\begin{figure}
  \centering
  \begin{tabular}{cc}
      \includegraphics[height=3.5cm]{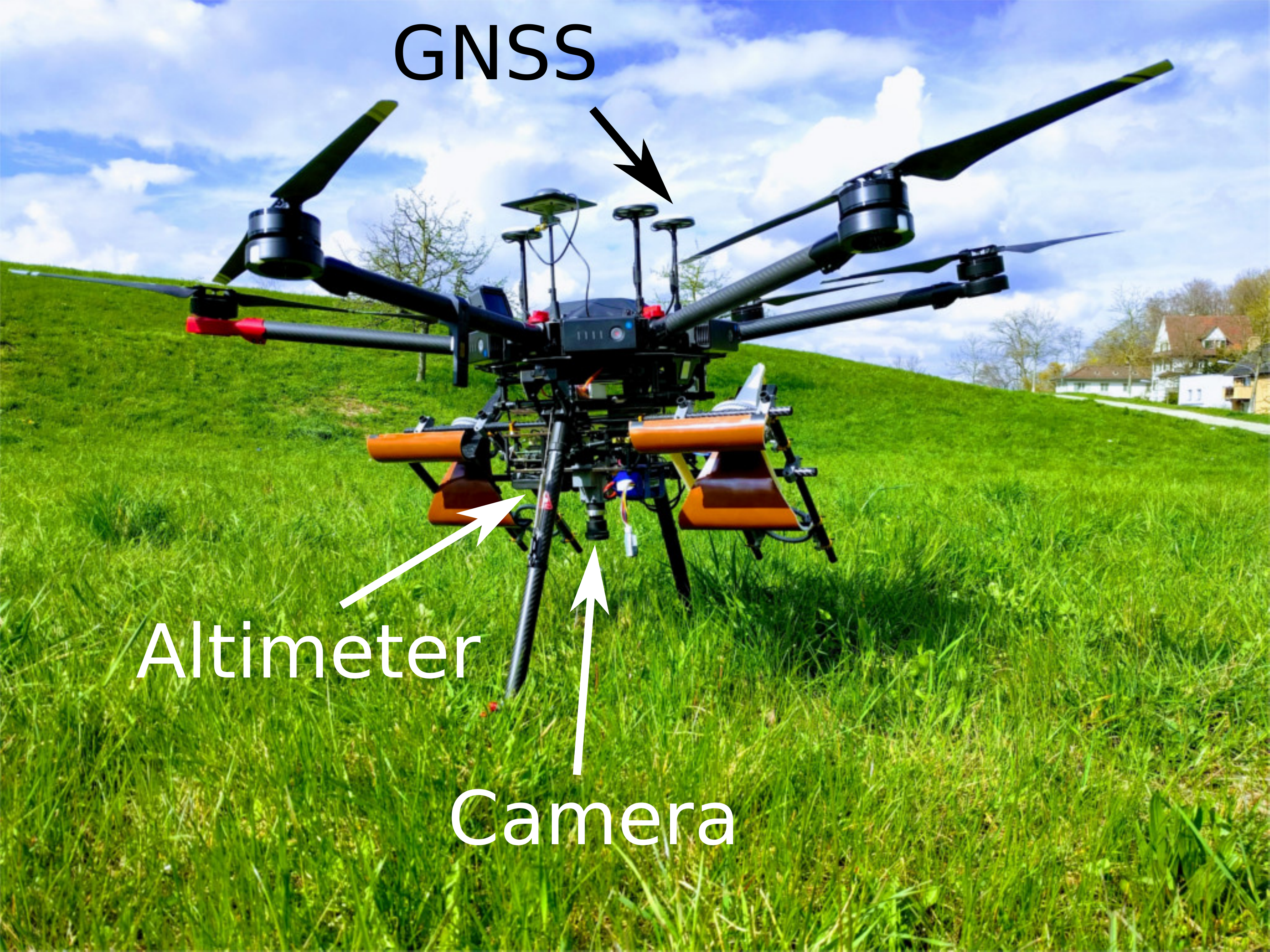} &
      \includegraphics[height=3.5cm]{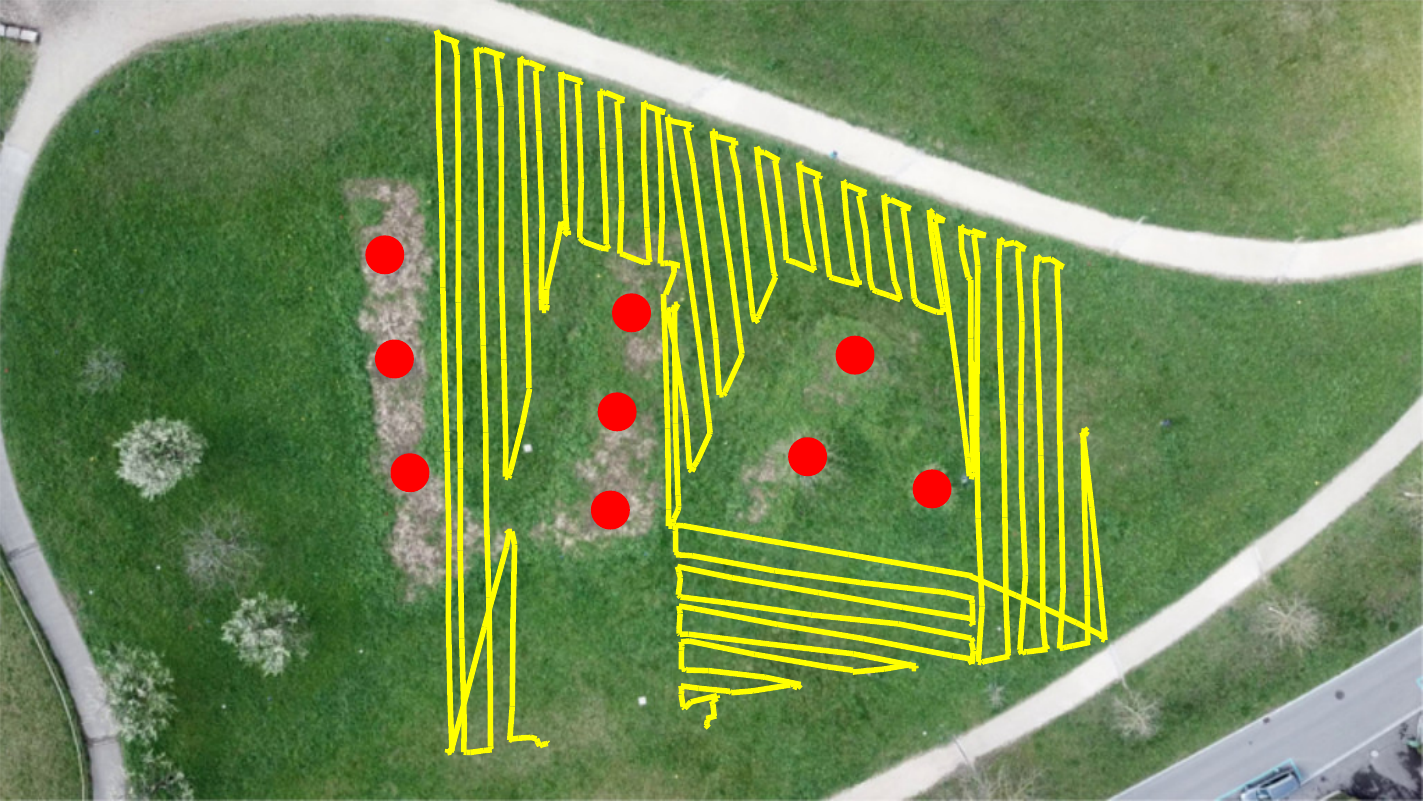} \\[6pt]
      (a) The \ac{MAV} platform.  &
      (b) Coverage path over sloped terrain (trees in red). 
\end{tabular}
  \caption{Deployment of our landmine detecting \ac{MAV}.
  The low-altitude coverage path was generated in the field.
  The operator defined obstacle boundaries to avoid trees and the planner found an optimal path. Video: \url{https://youtu.be/u1UOqdJoK9s}}
  \label{fig:eyecatcher}
\end{figure}

\Acp{MAV} such as the DJI M600 shown in \autoref{fig:eyecatcher} present ideal platforms for supporting demining efforts and other critical remote sensing tasks. 
These commercially available platforms are fast to deploy and can collect useful data that are often inaccessible to other modes of sensing \cite{pajares2015overview}. 
For demining applications, this means that \iac{MAV} carrying \iac{SAR} system can be flown over a minefield to collect in-situ, high precision measurements for locating buried landmines \cite{schartel2018uav}.

At present, the target coverage area is \SI{1}{\hectare} per battery charge.
Furthermore, since the \ac{MAV} must fly at relatively low altitudes to emit enough energy into the ground, structures in the environment such as trees or buildings can force \acp{NFZ} across the space. 
These obstacles can be especially problematic since they are often omitted from maps and must be dealt with at deployment time. 
This results in a need for an autonomous coverage planning and execution solution that can rapidly replan trajectories in the field.

Many existing commercial mission planners, e.g., Ardupilot Mission Planner\footnote{\url{http://ardupilot.org/planner/}}, Pix4DCapture\footnote{\url{https://www.pix4d.com/product/pix4dcapture}}, Drone Harmony\footnote{\url{https://droneharmony.com}}, and
DJIFlightPlanner\footnote{\url{https://www.djiflightplanner.com/}}
provide implementations of 2D coverage planners which allow specifying polygon flight corridors and generating lawnmower patterns. 
However, these mission planners aim at high-altitude, open space, top-down photography and thus neither allow specifying obstacles within the polygon nor consider the polygon edges as strict boundaries. 
Furthermore, due to safety limitations of commercial platforms, most commercial planners enforce a minimum altitude which is above the required \SI{3}{\metre} of our \ac{SAR} system.

While solutions for generating coverage paths in general polygon environments exist, the implementations are either not publicly available, computationally unsuitable for complex large environments or lack the full system integration that would allow a user to freely designate the flight zone.

In this work we develop a complete pipeline for rapidly generating 2D coverage plans that can account for \acp{NFZ}. 
Our proposed solution takes as input a general polygon (potentially containing \acp{NFZ}) and performs an exact cell decomposition over the region. 
Initial sweep patterns are computed for each cell from which feasible flight trajectories are computed. 
Finally, the \ac{E-GTSP} across the complete cell adjacency graph is solved to minimize the total path time.

Results comparing our proposed coverage planner to exhaustively searching a coverage adjacency graph demonstrate an order of magnitude speedup in the computation time for each trajectory. Specifically, we show that our solution's computation time grows reasonably with increasing map complexity, while the exact solution grows unbounded. Furthermore, the flight path computed by our method is \SI{14}{\percent} shorter than those found by solving a regular \ac{TSP} while providing the same level of coverage.

The contributions of this work are:
\begin{itemize}
  \item \Iac{E-GTSP} based fast 2D coverage planning algorithm that allows specifying polygonal flight zones and obstacles.
  \item Benchmarks against existing solutions.
  \item An open source \ac{ROS} implementation.\footnote{\url{https://github.com/ethz-asl/polygon_coverage_planning}}
\end{itemize}

We organize the remainder of the paper as follows.
Section~\ref{sec:related_work} presents related work.
In \autoref{sec:computational_geometry} we present the background in computational geometry to generate the sweep permutations. 
In \autoref{sec:gtsp} we present our coverage planner formulation to solve for the optimal coverage pattern. 
Finally, we validate our method in \autoref{sec:results} before we close with concluding remarks in \autoref{sec:conclusions}.

\section{Related Work}
\label{sec:related_work}
Coverage planning in partially known environments is an omnipresent problem in robotics and includes applications such as agriculture, photogrammetry and search. It is closely related to the \ac{AGP} and \ac{TSP}, which are NP-hard \cite{o1983some,papadimitriou1977euclidean}, and for which heuristic solutions have been developed to cope with high dimensional problems. The coverage planning problem can be stated as: given a specified region and sensor model, generate a plan for a mobile robot that provides complete coverage, respects motion and spatial constraints, and minimizes some cost metric (often path length or time). The underlying algorithms to solve sweep planning in a 2D polygon map are usually either based on approximate or exact cellular decomposition. The state of the art in robotic coverage planning is summarized in \cite{galceran2013survey,cabreira2019survey}.

Early work in robot coverage planning developed approximate cellular decomposition methods. In \cite{zelinsky1993planning}, the target region is first decomposed into connected uniform grid cells such that coverage is achieved by visiting all cells. The authors presented a coverage wave front algorithm that obeyed the given starting and goal cells. Unfortunately, grid based methods grow linearly in memory with the area to be covered and exponentially in finding an optimal path, making them unsuitable for large areas with small sensor footprints. Furthermore, guiding the wave front algorithm to pick a route that meets all mission requirements involves designing and tuning a specific cost function for each application. Spanning-trees are an alternative method to solve this problem \cite{gabriely2001spanning}. However, by default those result in a rather large number of turns which is undesirable for \acp{MAV}. 

A simple approach stems from the idea that a polygonal region can be covered using sequential parallel back-and-forth motions, i.e. lawnmower or boustrophedon paths. This approach for 2D top-down lawnmower patterns in polygonal maps dominates commercial aerial remote sensing. The user defines a polygonal target region and a specified sweep direction, and the planner generates a path consisting of a sequence of equally-spaced sweep lines connected by turns. The generated coverage paths are time-efficient, complete, and intuitive to the user. However, these planners cannot account for \acp{NFZ} nor do they satisfy our targeted flight altitude requirements. Furthermore, except for Ardupilot Mission Planner, these commercial systems do not allow modification as they are closed source.

Exact cellular decomposition is a geometric approach that can handle \acp{NFZ} by dividing a configuration space into simpler component cells, whose union is the complete free space. A full coverage path can then be found by creating a boustrophedon path in each cell and connecting all cells. The individual cell coverage plans are connected in an adjacency graph and solving the resulting \ac{TSP} solves the coverage problem. Compared to grid-based methods, exact decomposition generally results in significantly fewer cells in simple large environments and thus a smaller \ac{TSP}. Choset and Pignon \cite{choset1998coverage} present the standard solution in 2D environments, which has been adapted by many planners for robot coverage \cite{galceran2015coverage,bahnemann2017decentralized}. Several improvements have also been proposed to minimize the number of cell traversals and the number of turns along the coverage path. Namely, methods for optimizing over the sweep line direction rather than using a fixed direction for all cells have been proposed \cite{li2011coverage, torres2016coverage}. However, these come at the cost of increased complexity when solving for the optimal path through the cell-connectivity graph.

In our work we revisit the exact cellular decomposition method and also generate multiple possible sweep directions per cell, however we formulate the resulting search as \iac{E-GTSP}, allowing us to use a state of the art genetic solver that handles significantly larger problem sizes. The \ac{E-GTSP}, also known as \ac{TSP} with neighborhoods, is a generalization of the classical \ac{TSP} \cite{mitchell2000geometric}. The goal is to find a shortest tour that visits exactly one node in each of a set of $k$ neighborhoods. As first proposed by Waanders \cite{waanders2011coverage}, we cluster all possible sweep patterns of a cell in a neighborhood and search for the shortest path that includes exactly one sweep pattern per cell.

Bochkarev and Lewis \cite{bochkarev2016minimizing,lewis2017semi} took this \ac{E-GTSP} formulation even further. Instead of using per-cell predefined sweep patterns, they precompute a set of globally good sweeping directions for each monotone cell and build a graph over all individual straight segments where a neighborhood is defined by traversing a segment in either direction. Compared to our approach their approach can give better solutions where the robot traverses from one cell to another cell without covering it completely first at the expense of a more complex \ac{E-GTSP} and a predefined sweep direction.

To solve the \ac{E-GTSP}, several options exist. Exact solvers like \cite{fischetti1997branch} only work reliably for small problem sizes. Converting the problem into a directed graph using a product graph \cite{rice2012exact} and solving it with an optimal graph solver, e.g. Dijkstra, basically falls back to solving it exhaustively as in \cite{choset1998coverage} and \cite{torres2016coverage}, which remains intractable for larger problem sizes. Thus a practical solution for our problem sizes is to use heuristic solvers. Helsgaun \cite{helsgaun2015solving} transforms the \ac{E-GTSP} into a \ac{TSP} and uses an approximate \ac{TSP} solver. The \ac{GK}, on the other hand, is a memetic algorithm that approximately solves the \ac{E-GTSP} directly and is faster but with reduced performance for large problems \cite{gutin2010memetic}. 

\section{Geometric Path Generation}
\label{sec:computational_geometry}
\begin{figure*}
    \null\hfill
    \subfloat[Input \ac{PWH}]{\includegraphics[width=0.3\textwidth]{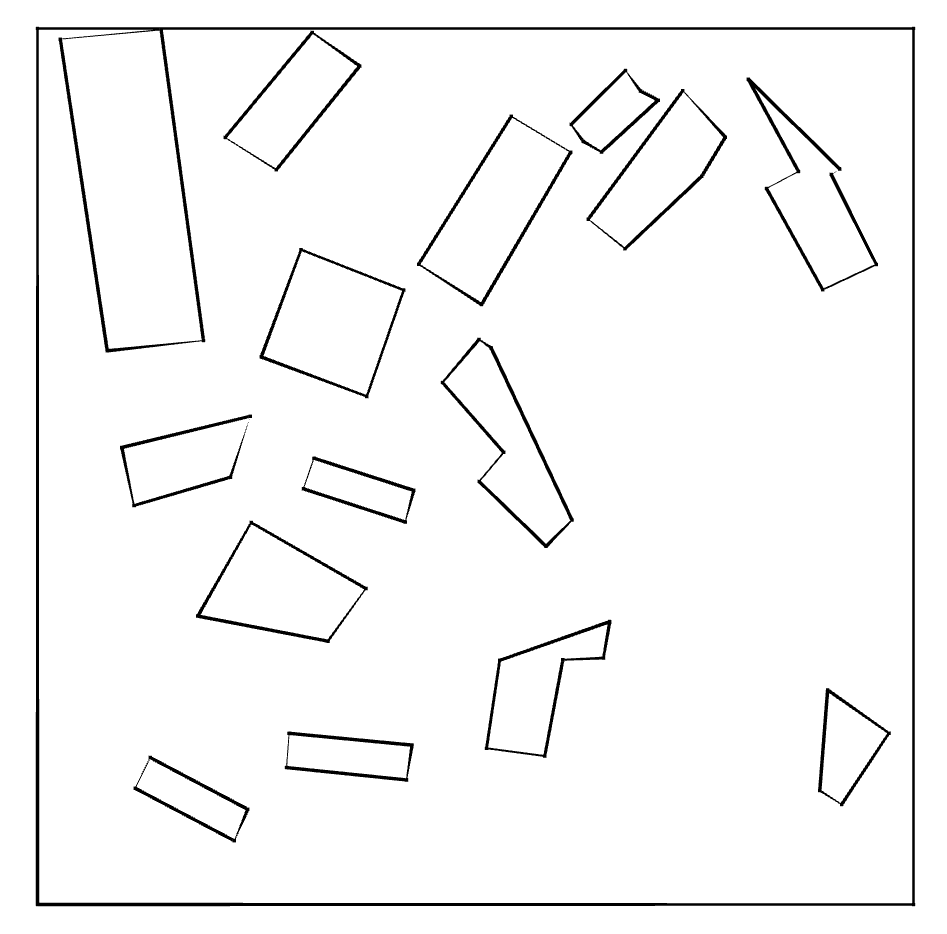}}
    \hfill
    \subfloat[Cell decomposition]{\includegraphics[width=0.3\textwidth]{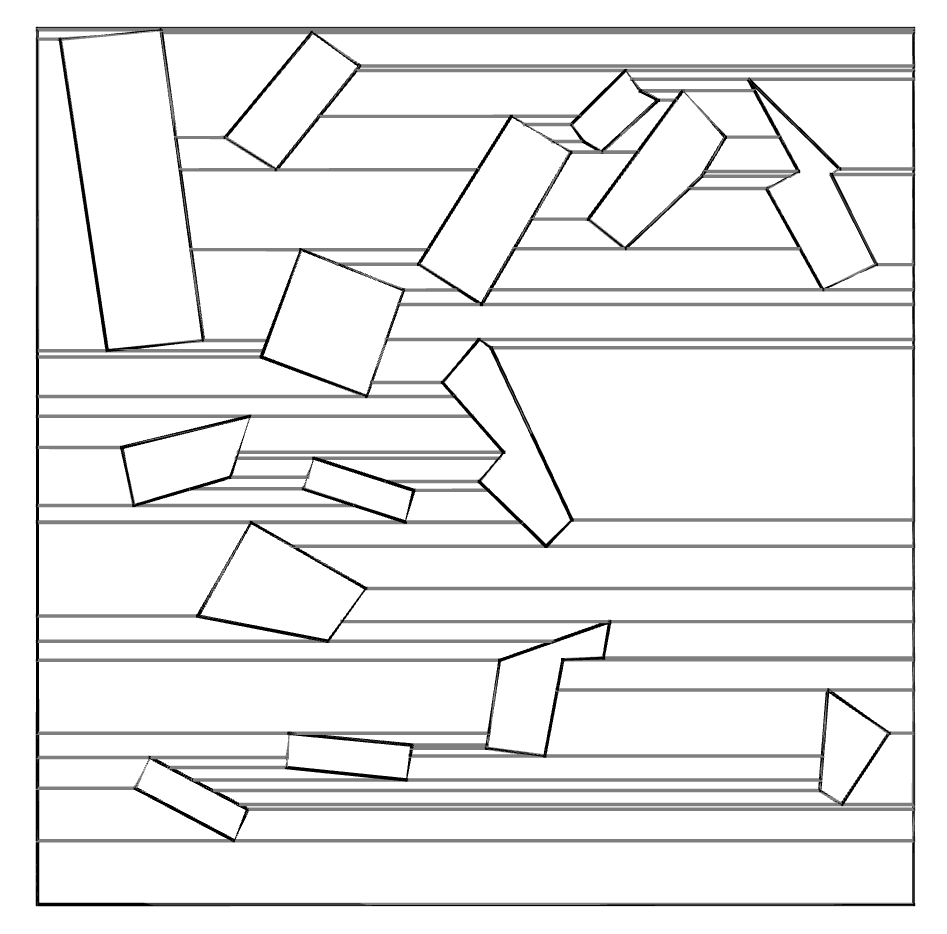}}
    \hfill
    \subfloat[Optimal sweep pattern]{\includegraphics[width=0.3\textwidth]{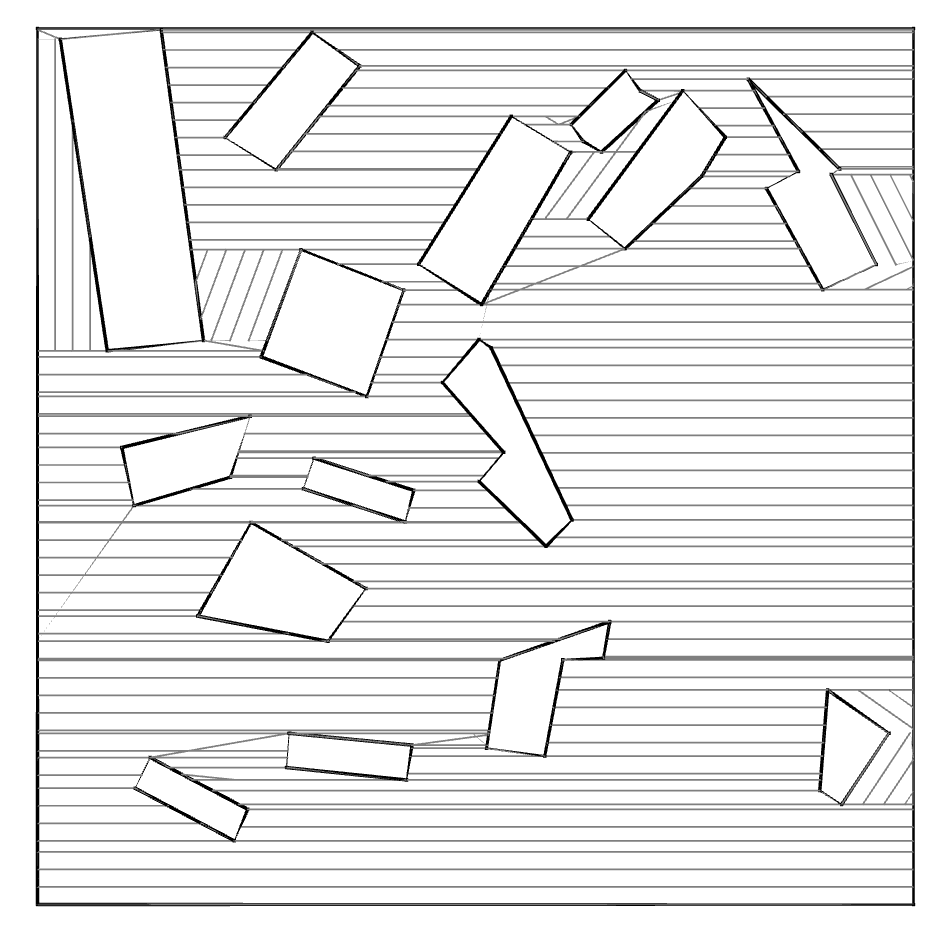}}
    \hfill\null
    \caption{The coverage algorithm on a synthetic map with $15$ obstacles and $52$ hole vertices.}
    \label{fig:pipeline}
\end{figure*}

In order to solve the coverage path problem we follow the route of exact cellular decomposition techniques outlined in \autoref{fig:pipeline}.
We decompose a \ac{PWH} into cells, in a way that guarantees that each cell can be fully covered by simple boustrophedon paths.
Our algorithm creates a permutation of sweeping directions for each cell and finds a shortest route that connects and covers every cell to define the \textit{coverage path}.

\subsection{Sweep Pattern Permutation}
A \textit{sweep pattern} describes the combination of \textit{straight segments} and \textit{transition segments} that covers a single polygon cell.
A continuous parallel sweep pattern can be generated perpendicular to any monotone direction of a simple polygon.
Without loss of generality we can consider the monotone direction the $y$-direction of the polygon.
A polygon $P$ is considered $y$-monotone, if any line in the $x$-direction intersects $P$ at a single point, a segment or not at all.
In other words the line intersects $P$ at most twice \cite{de1997computational}.

\begin{figure}
    \centering
    \begin{tabular}{cc}
    \includegraphics[height=1.0in]{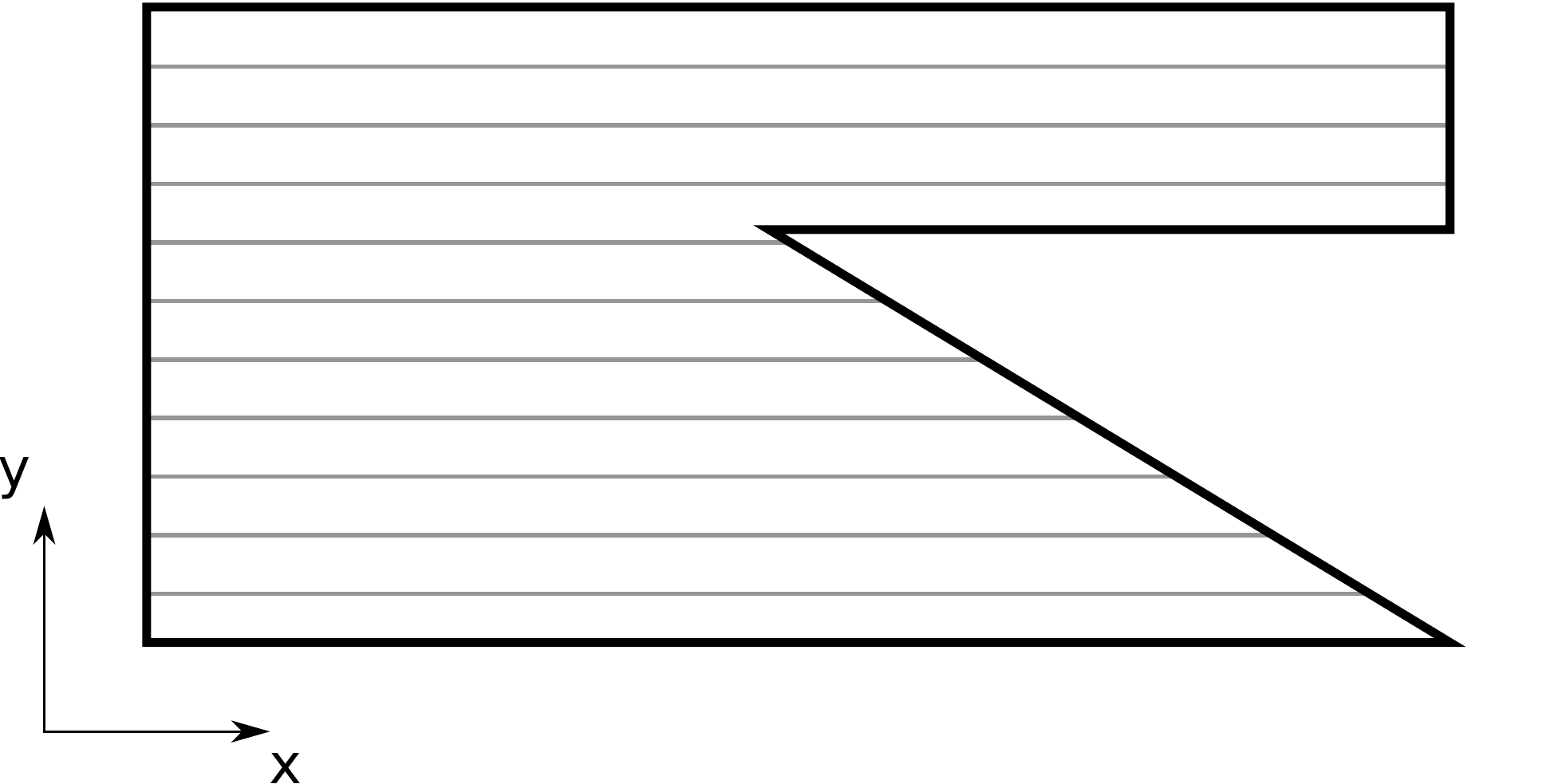} &
    \includegraphics[height=1.0in]{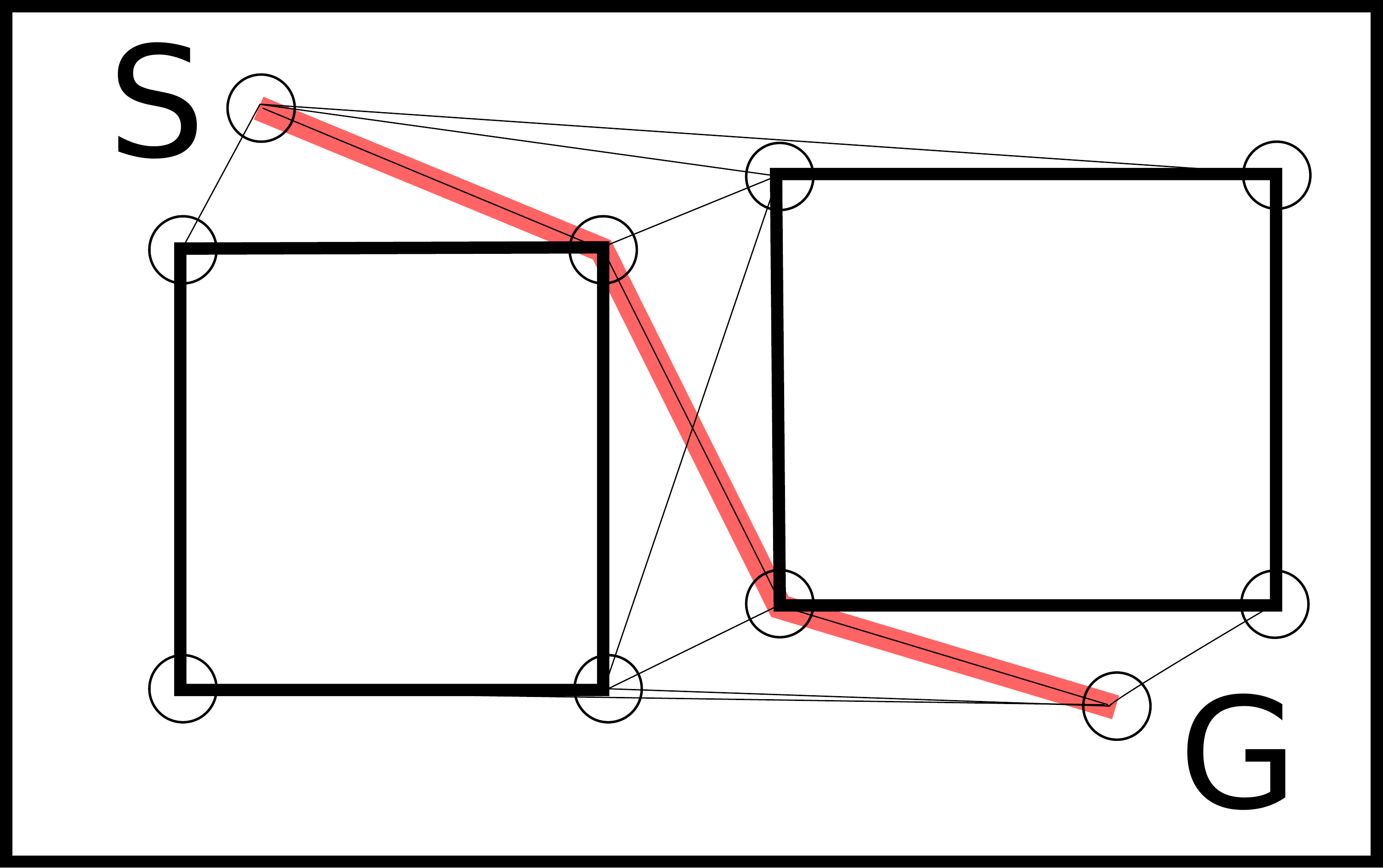} \\
    (a) Straight segments in a $y$-monotone polygon. &
    (b) Shortest path transition segments.
    \end{tabular}
    \caption{A sweep pattern consists of a set of parallel offsetted straight segments sequentially connected via transition segments. The straight segments are generated along edges that are perpendicular to a monotone direction (left). The transition segments are Euclidean shortest paths along the reduced visibility graph (right).}
    \label{fig:shortest_path}
    \label{fig:monotone}
\end{figure}

\autoref{fig:monotone} (a) shows the straight segment generation in a $y$-monotone polygon.
We initialize the first straight segment at the bottommost vertex parallel to the $x$-axis, which we refer to as \textit{sweep direction}.
In general, we restrict the sweep directions to be collinear to one of the edges of our polygon, as these directions have been proven to lead to a minimum number of straight segments to cover the polygon \cite{huang2001optimal}. 
The individual straight segments are generated by alternating between intersecting a line in the $x$-direction with the polygon and offsetting the line from the bottommost towards the topmost vertex.
The sensor footprint hereby defines the offset distance.

\begin{figure}
    \centering
    \begin{tabular}{cccc}
          \includegraphics[height=0.55in]{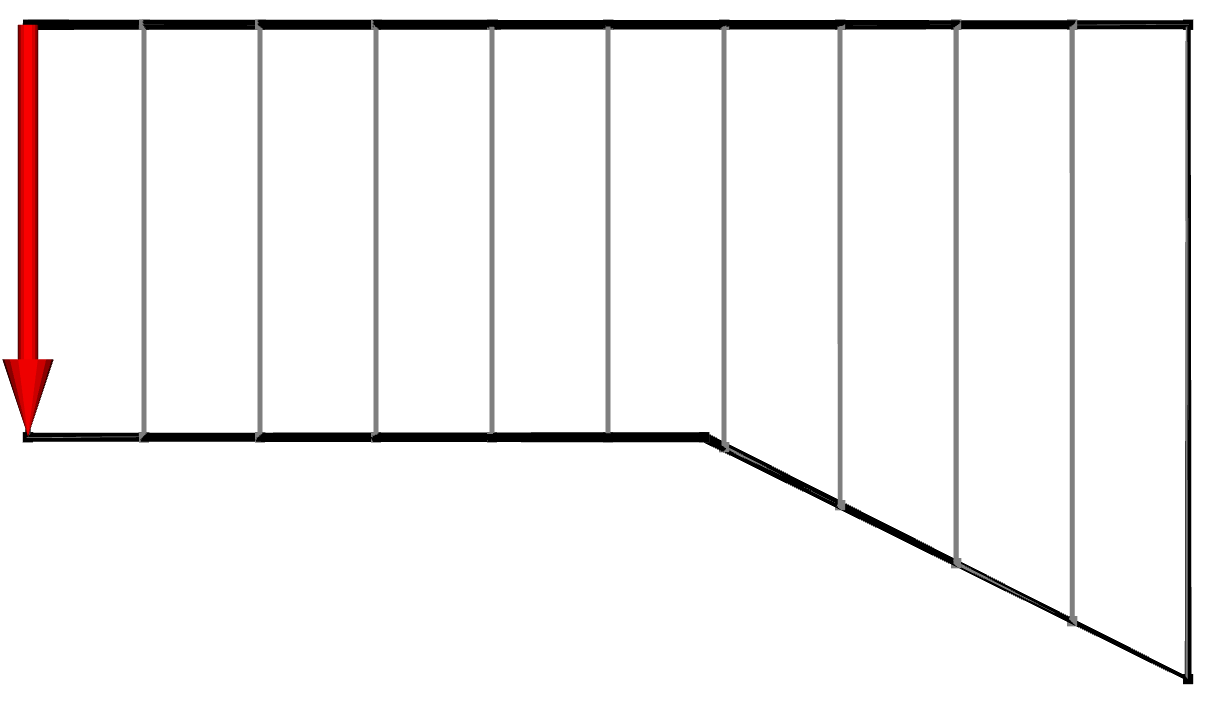} &  
          \includegraphics[height=0.55in]{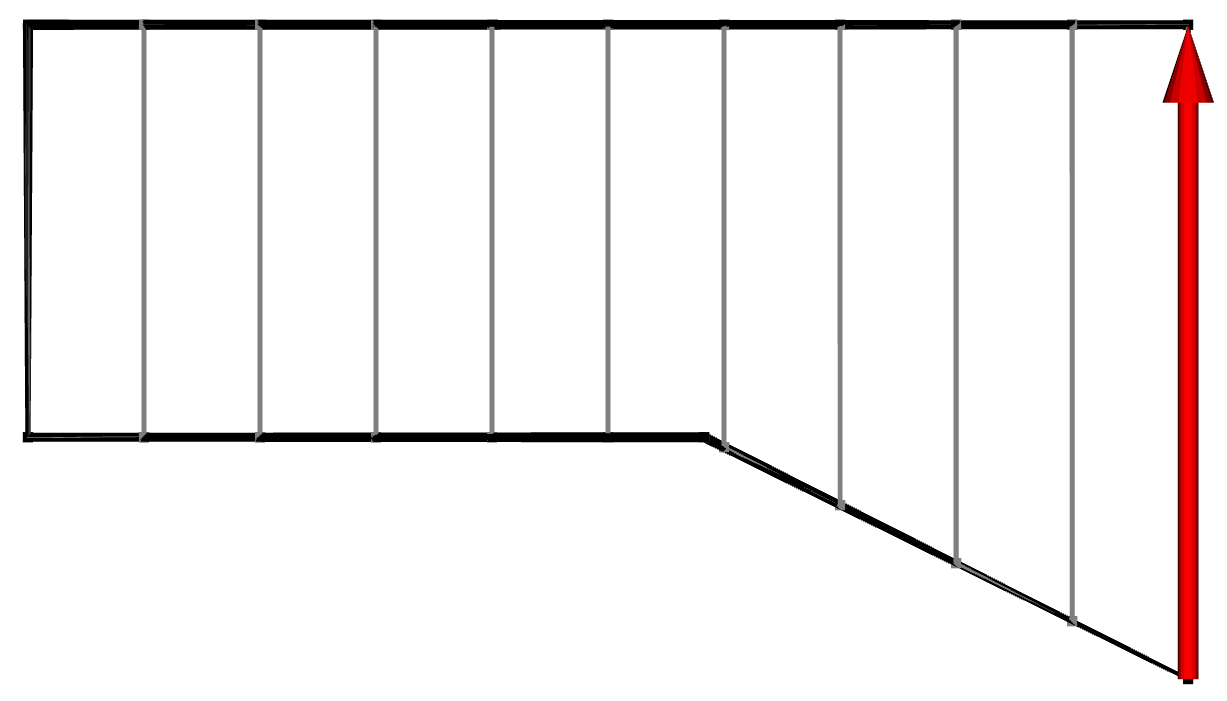} &  
          \includegraphics[height=0.55in]{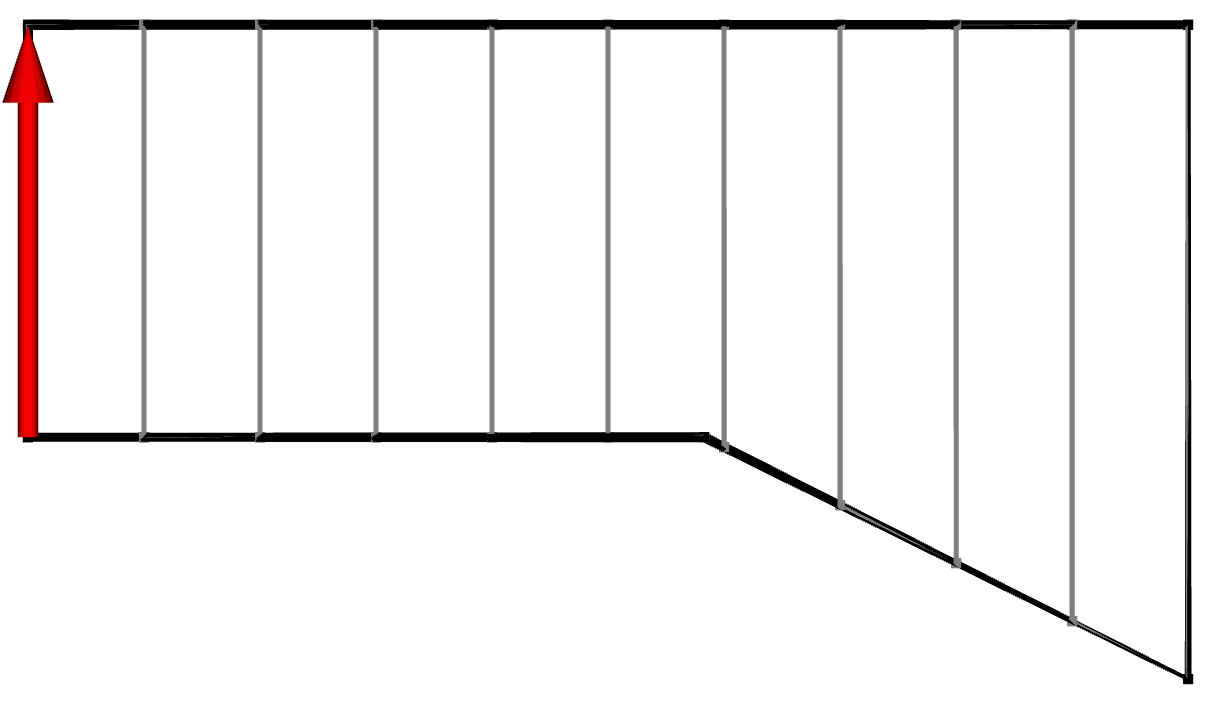} &
          \includegraphics[height=0.55in]{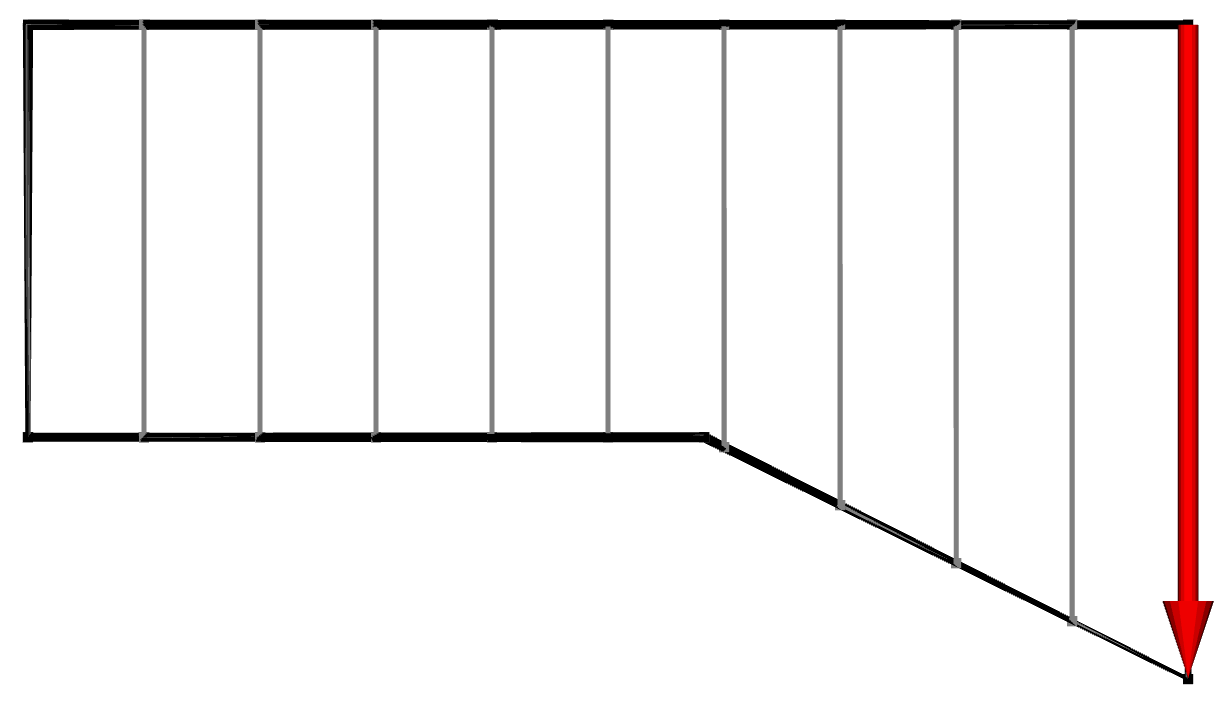} \\[6pt] 
          \includegraphics[height=0.55in]{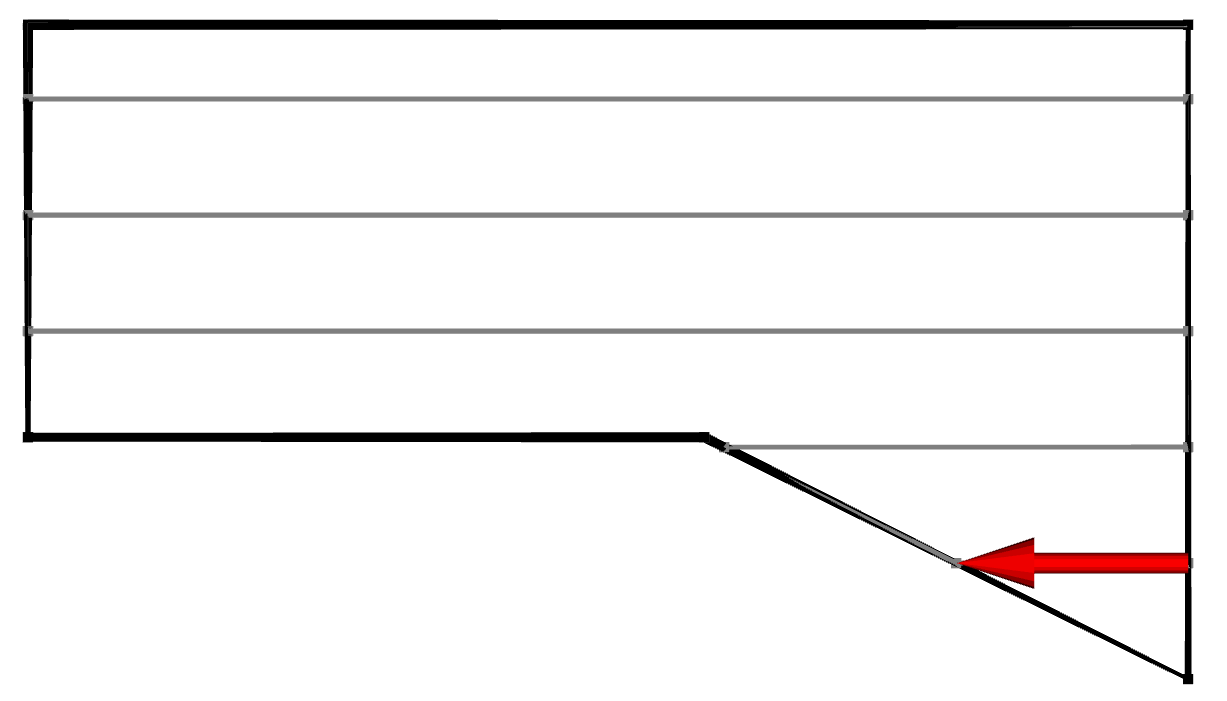} &  
          \includegraphics[height=0.55in]{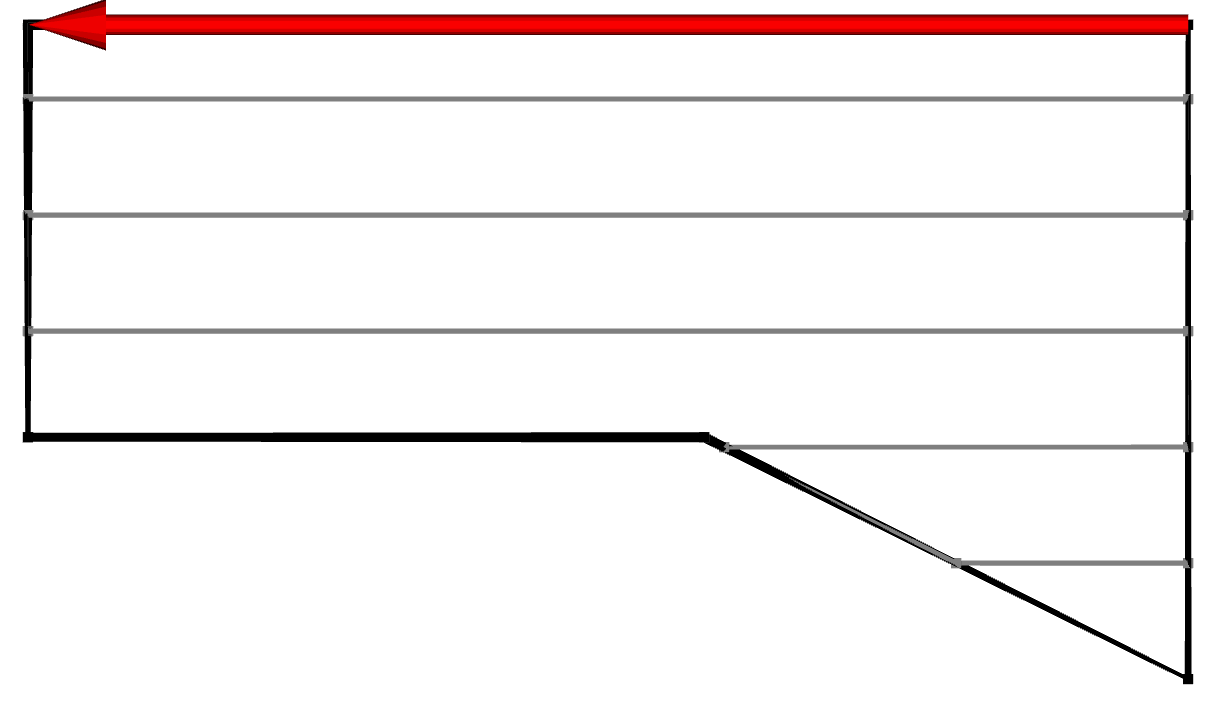} &
          \includegraphics[height=0.55in]{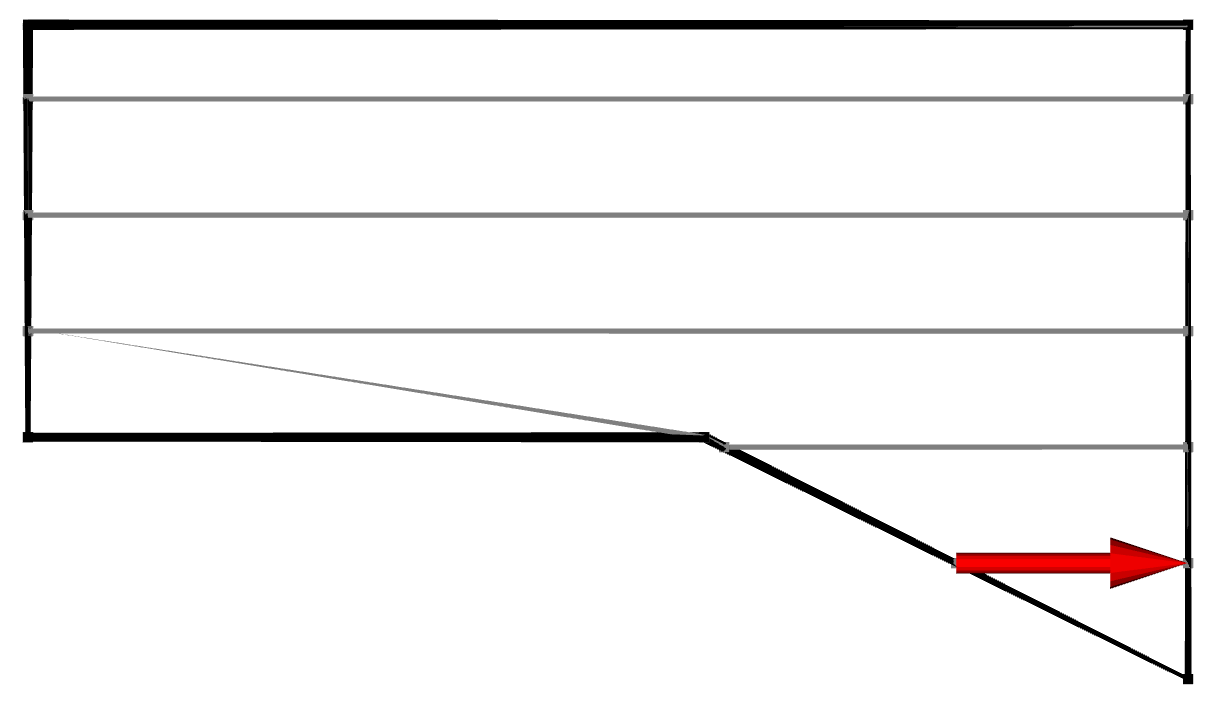} &  
          \includegraphics[height=0.55in]{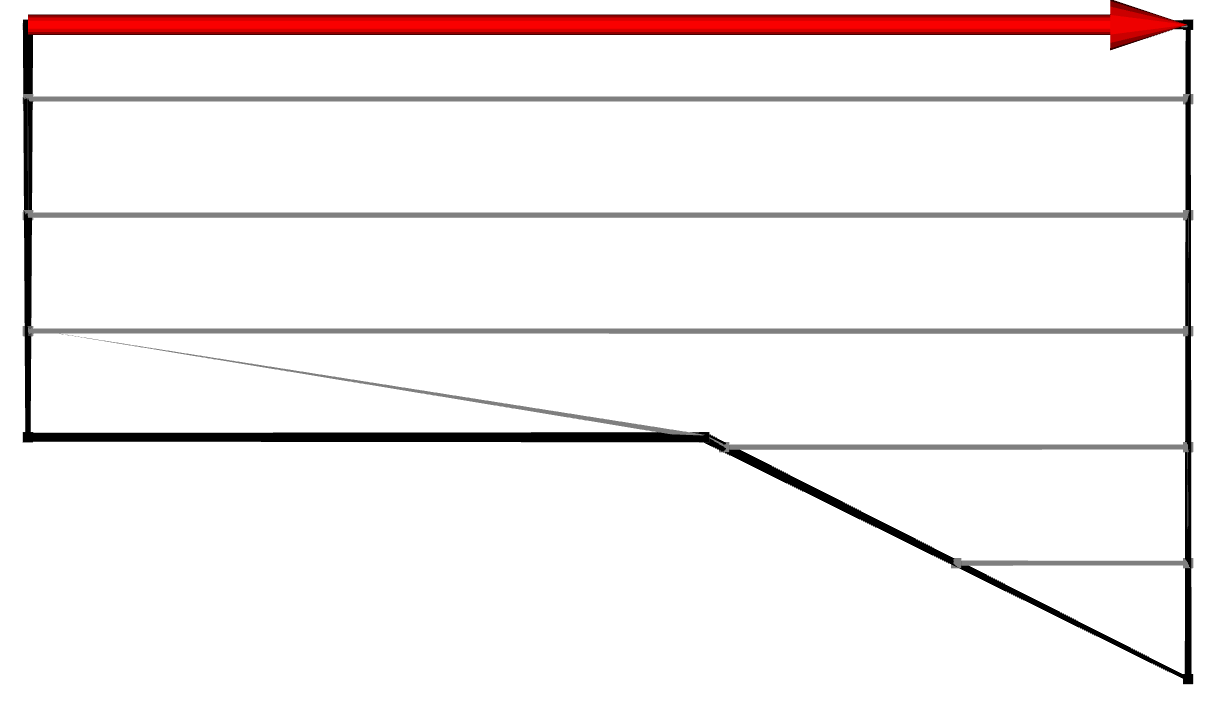} \\[6pt] 
          \includegraphics[height=0.55in]{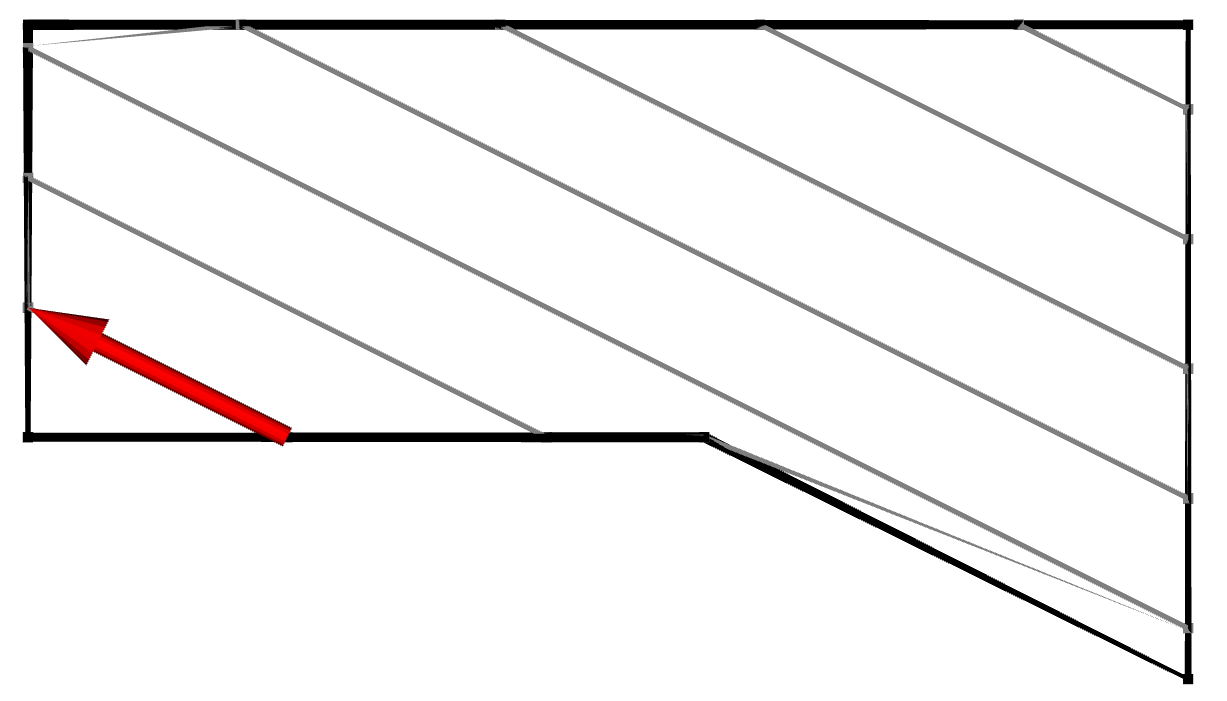} &
          \includegraphics[height=0.55in]{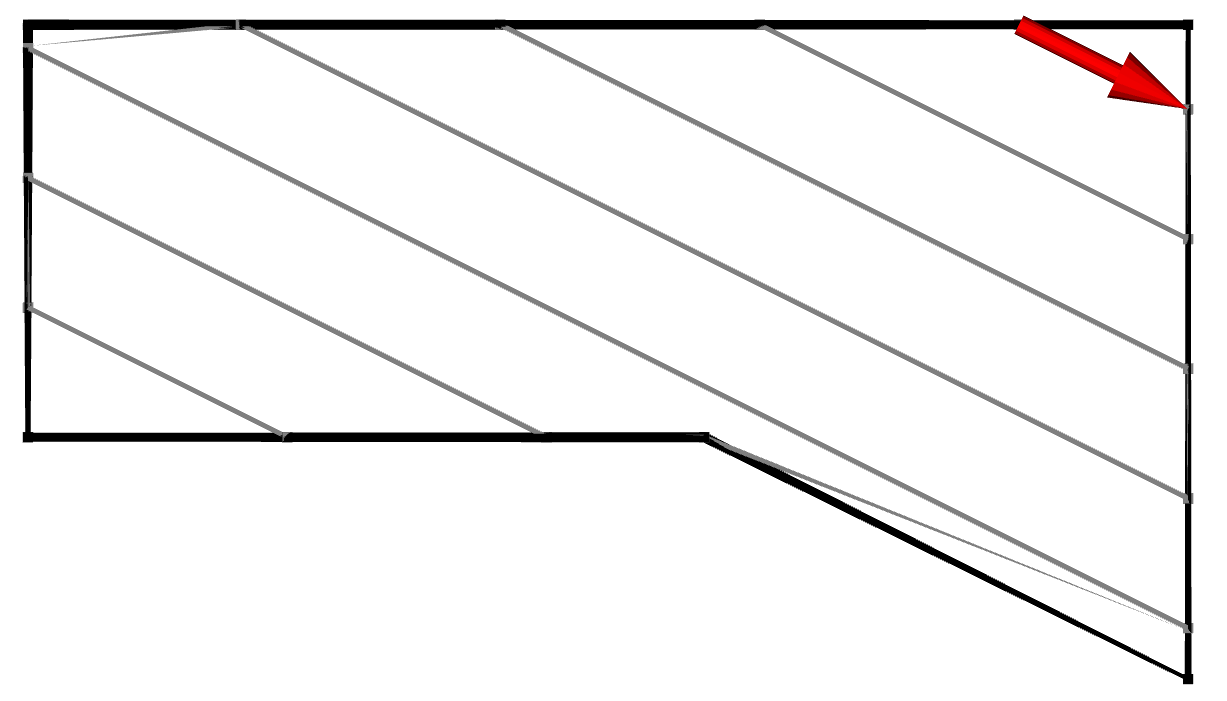} &  
          \includegraphics[height=0.55in]{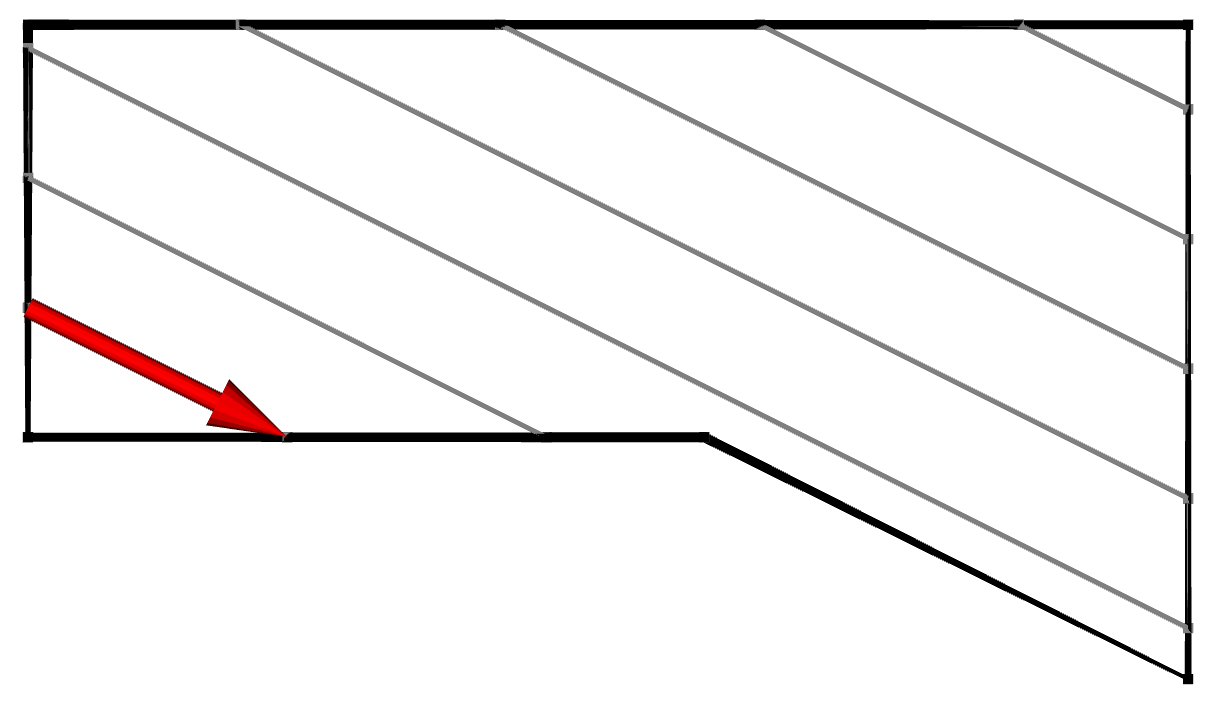} &  
          \includegraphics[height=0.55in]{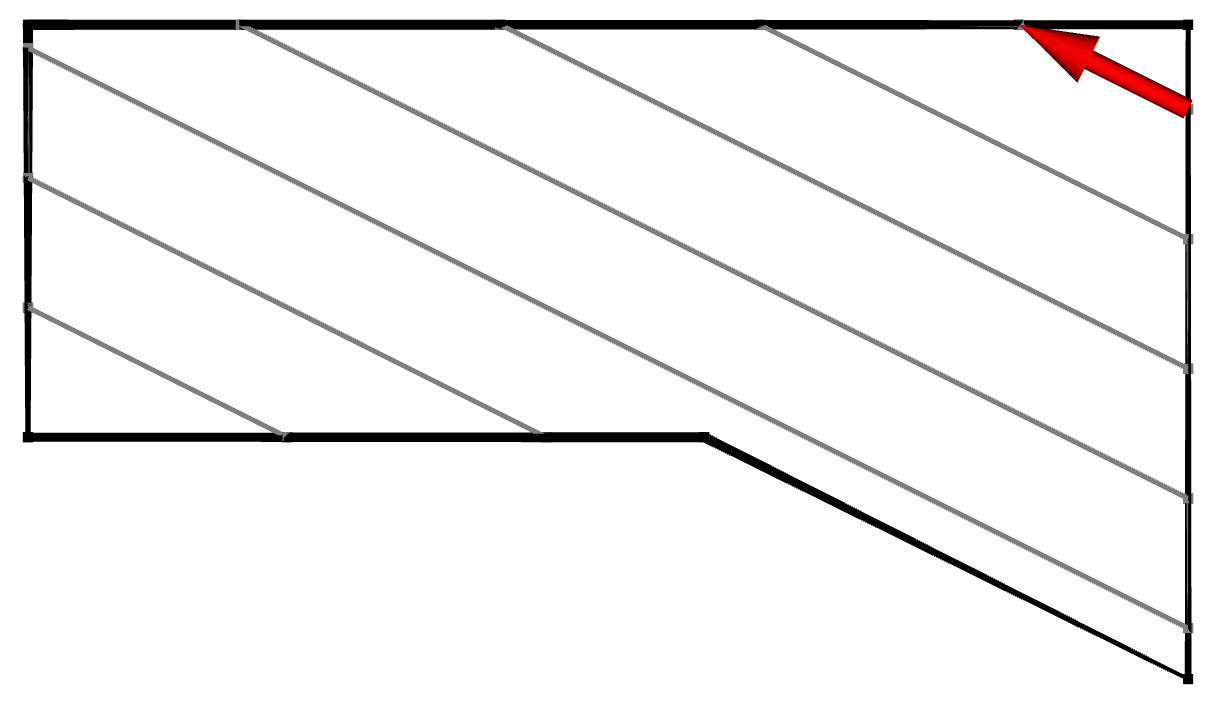}
    \end{tabular}
    \caption{Each \textit{sweepable} direction (rows) has four \textit{sweep permutations} (columns) based on the start vertex and start direction (red arrow).}
    \label{fig:permutation}
\end{figure}

Based on this construction criterion, the straight segments can start clockwise or counter-clockwise at the bottom vertex and go from the bottom to the top or from top to bottom.
Thus we generate four possible sweep patterns per \textit{sweepable} direction as shown in \autoref{fig:permutation}.

To connect two straight segments (and later to connect two sweep patterns) we calculate the Euclidean shortest path that avoids collision with the \ac{NFZ}.
The Euclidean shortest path in a \ac{PWH} is computed along the \textit{reduced visibility graph} \cite{latombe2012robot}.
\autoref{fig:shortest_path} (b) shows an example solution (red) using A$^{*}$ to search the graph \cite{hart1968formal}.
The graph node set (circles) consists of all non-convex hull vertices and convex hole vertices.

\subsection{Polygon Decomposition}
\begin{figure}
    \centering
    \begin{tabular}{llrr}
          \includegraphics[height=0.9in]{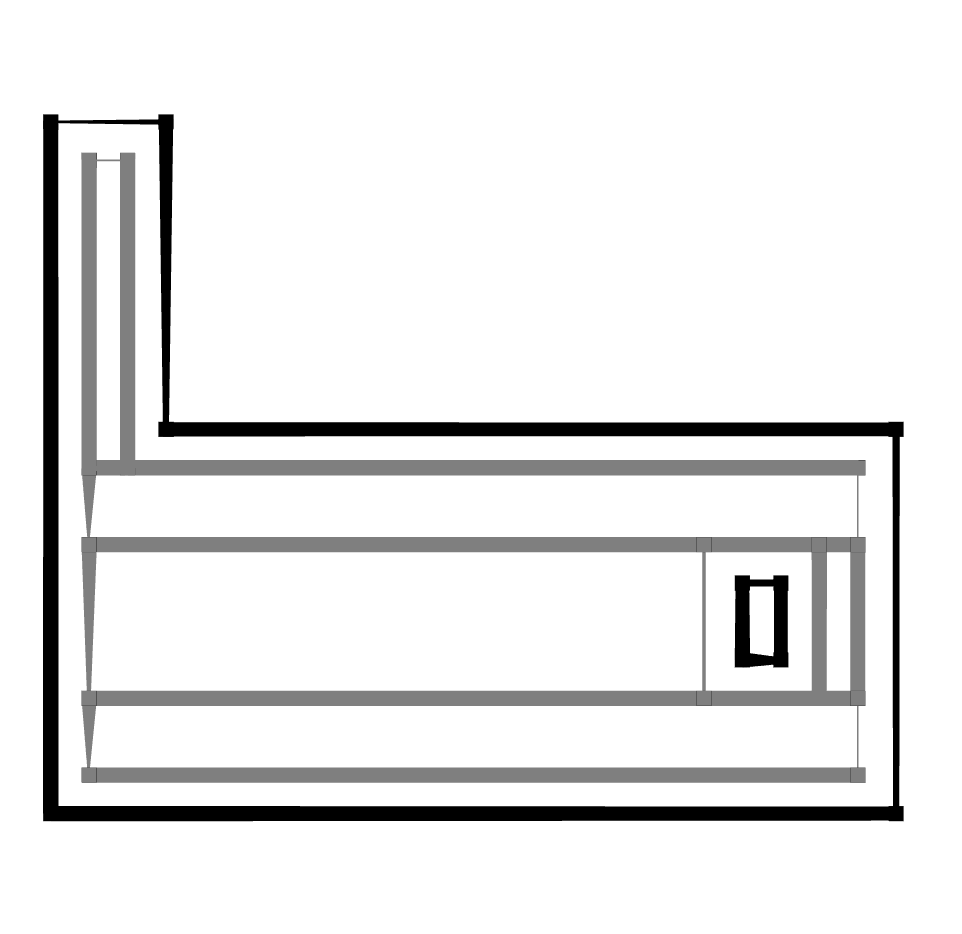} &
          \includegraphics[height=0.9in]{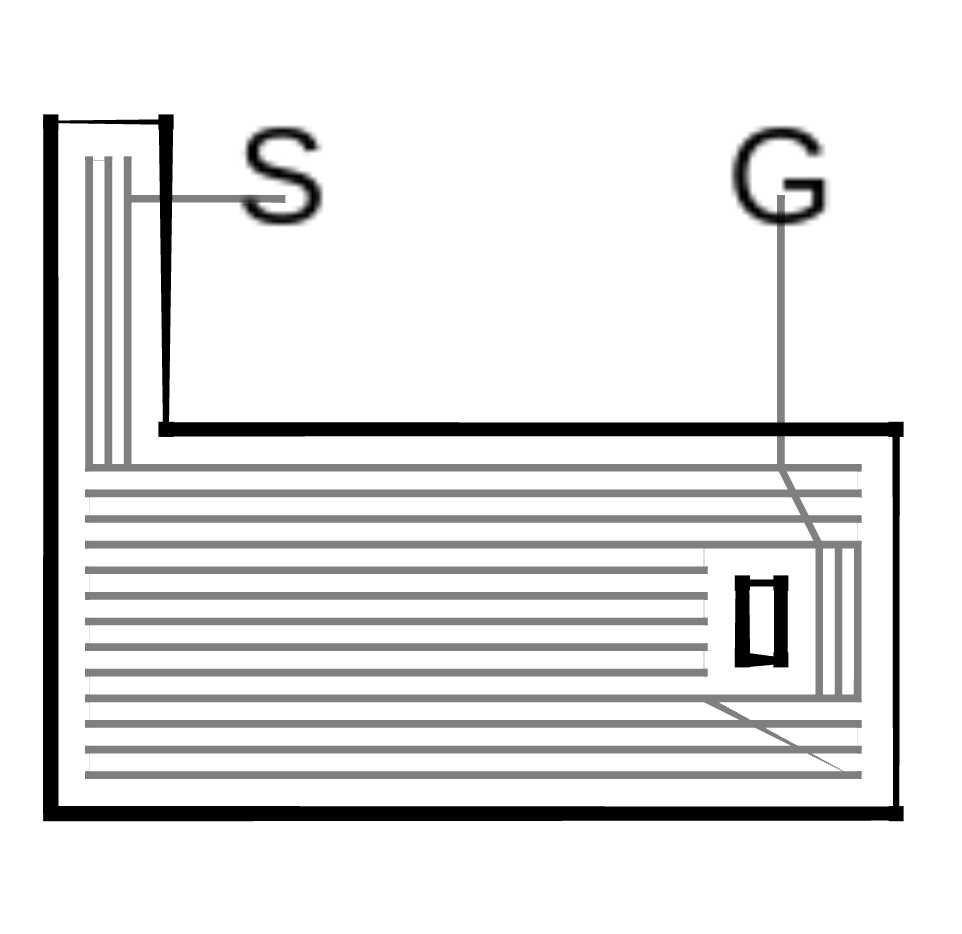} &  
          \includegraphics[height=0.9in]{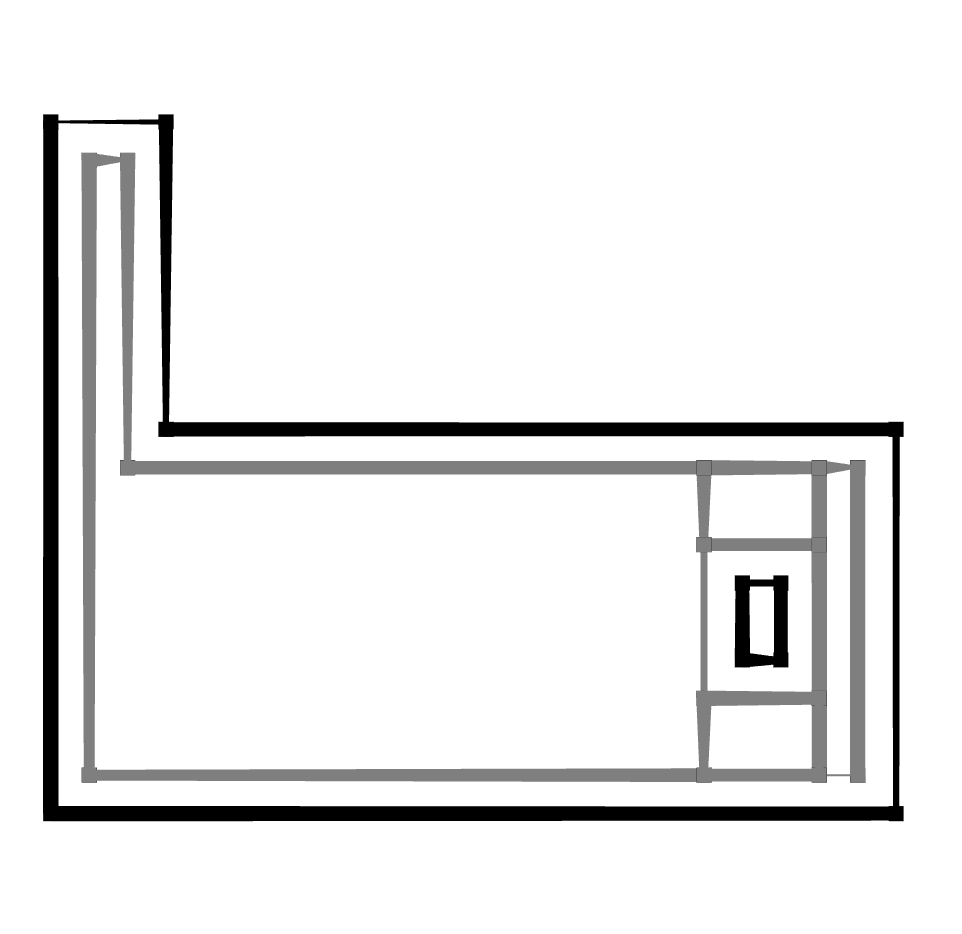} &
          \includegraphics[height=0.9in]{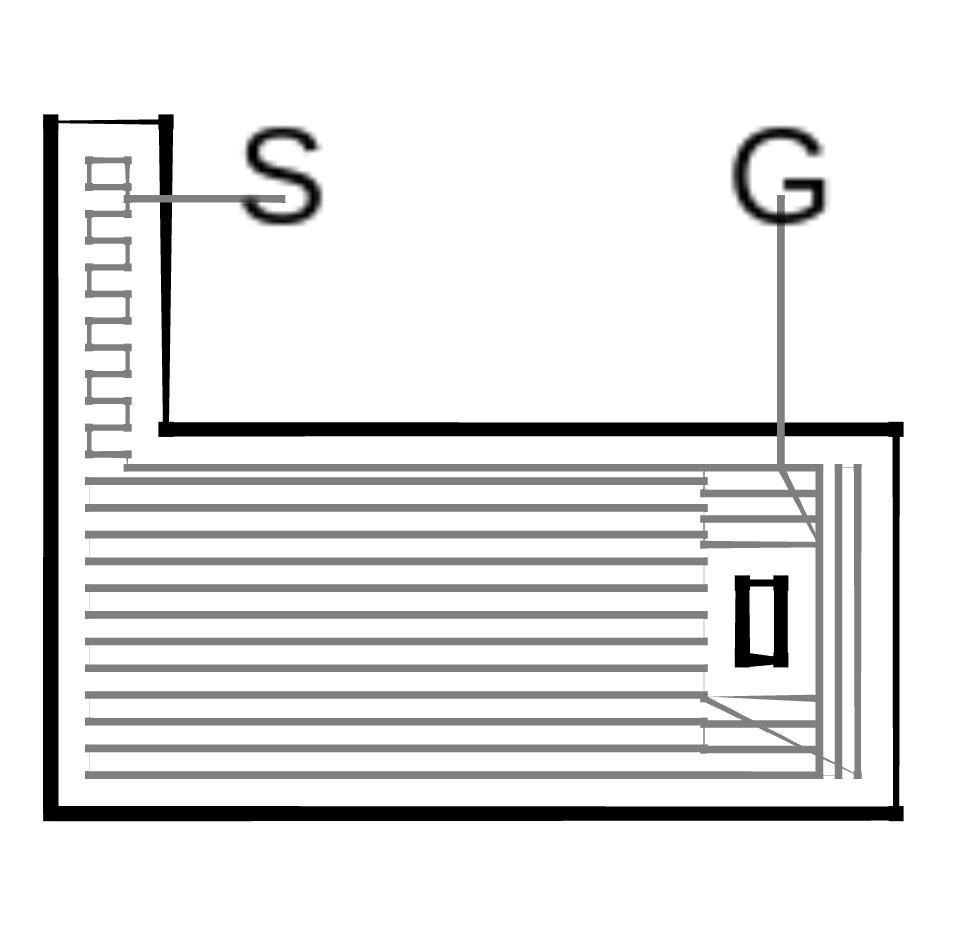} \\[6pt]
          \multicolumn{2}{c}{(a) \ac{TCD}} & \multicolumn{2}{c}{(b) \ac{BCD}} 
    \end{tabular}
    \caption{Qualitative comparison of \ac{TCD} and \ac{BCD}.
    While both adjust well to rectangular environments,
    the \ac{BCD} leads to fewer cells and redundant sweeps but can have degenerate sweeping behaviour in narrow regions.}
    \label{fig:decompositions}
\end{figure}

To cover a general \ac{PWH}, we partition it into monotone non-intercepting cells whose union is again the original \ac{PWH}.
In general any monotone decomposition would be feasible but we only consider the \ac{TCD} and the \acf{BCD} \cite{choset1998coverage,de1997computational}.
Both decompositions have different advantages as shown in \autoref{fig:decompositions}.
The \ac{TCD} provides a partitioning that adjusts well in rectangular scenes with multiple directions of extension.
The \ac{BCD} also adjusts well to rectangular scenes and usually leads to fewer cells and thus fewer redundant sweeps and traversal segments \cite{choset1998coverage}.
On the downside it can lead to degenerate sweeping behaviour in cells with narrow protrusions. 

Since both decompositions result from scan line algorithms, the scan line direction determines the set of cells.
To find a good decomposition direction we calculate the decomposition for every individual edge direction.
A potentially good decomposition is the decomposition with the smallest altitude sum $w$, where altitude refers to the monotone extension of a cell.
The minimum altitude sum corresponds to a minimum number of sweeps in the case of a fixed sweep direction \cite{huang2001optimal}. 
\begin{align}
    w &= \sum_{i=1}^m y_{\mathrm{max,}i} - y_{\mathrm{min,}i},
\end{align}
where $m$ is the number of monotone cells, $y_{\mathrm{max,}i}$ is the $y$-coordinate of the uppermost vertex and $y_{\mathrm{min,}i}$ is the $y$-coordinate of the lowermost vertex in a $y$-monotone polygon cell.

\section{Coverage Path Planning}
\label{sec:gtsp}
After decomposing the input \ac{PWH} into simple cells and generating a set of sweep patterns for each cell the planner has to find a shortest sequence of sweep patterns such that every cell, and thus the whole \ac{PWH}, is covered.
To solve this problem efficiently we formulate it as an \acl{E-GTSP}.

\begin{figure}
    \centering
    \includegraphics[height=1.5in]{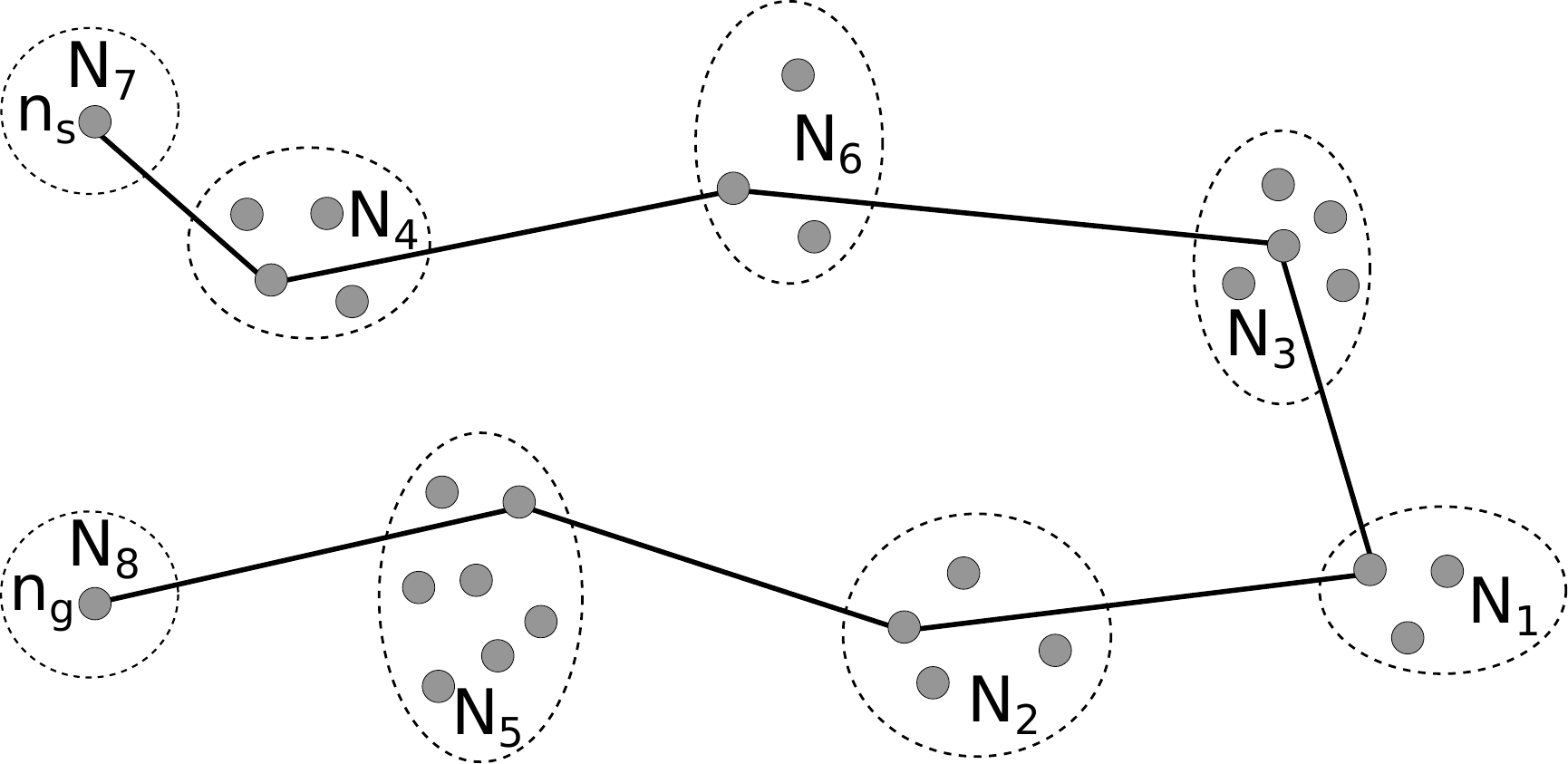}
    \caption{Visualization of the coverage planning \ac{E-GTSP}. The goal is to find the shortest path that visits exactly one sweep pattern (grey dot) in each polygon cell (dotted ellipsoid) while traveling from the start node $n_s$ to the goal node $n_g$.}
    \label{fig:gtsp}
\end{figure}

\autoref{fig:gtsp} sketches the adjacency graph $G = (N, A)$, where
$N = \{n_1 \ldots n_n\}$ is the node set and $A = \{ (n_i,n_j) : n_i, n_j \in N, i \neq j \}$ is the set of directed arcs.
The node set $N$ is divided into $m$ mutually exclusive and exhaustive clusters $N_1 \ldots N_m$, i.e., $N = N_1 \cup \ldots \cup N_m$ with $N_i \cap N_j = \emptyset ~ \forall i,j \in \{1 \ldots m\}, i \ne j $.

A node $n_i$ represents an individual sweep pattern that covers a single monotone polygon cell as shown in \autoref{fig:permutation} or the start or goal point.
Every monotone cell (and the start and goal point) represents an individual cluster $N_i$.
The arcs are the shortest path connecting the end of one sweep pattern with the start of the next sweep pattern in another cluster.
The start node has outgoing arcs to all nodes, and the goal node has incoming arcs from all nodes.

Every arc $(n_i,n_j)$ has a non-negative cost $c_{ij}$ which is the sum of the shortest path cost $t_{ij}$ from $n_i$ to $n_j$ and the cost $t_{j}$ to execute sweep pattern $n_j$.
\begin{align}
    c_{ij} = t_{ij} + t_{j}
\end{align}
Because the start and the end of a sweep pattern do not coincide, the cost matrix $\mat{C} = (c_{ij})$ is asymmetric, i.e., $c_{ij} \neq c_{ji}$.

\begin{figure}[t]
    \centering
    \begin{tabular}{ccc}
          \includegraphics[height=0.9in]{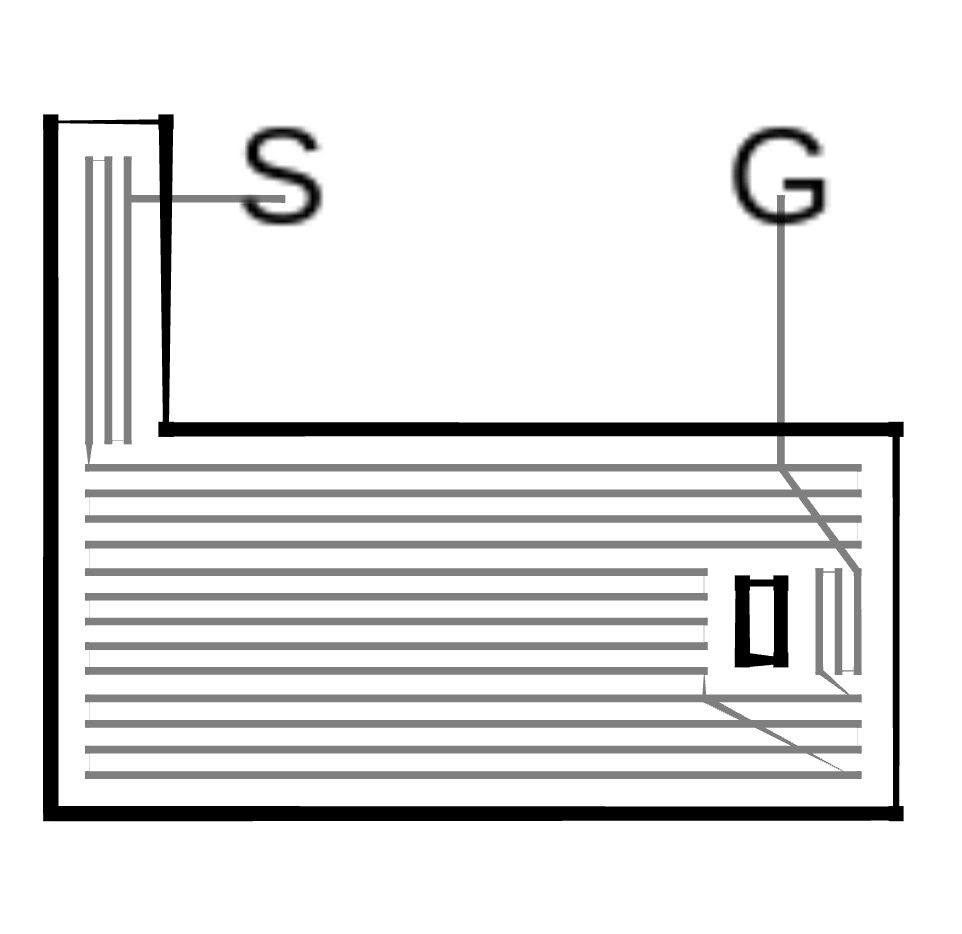} &  
          \includegraphics[height=0.9in]{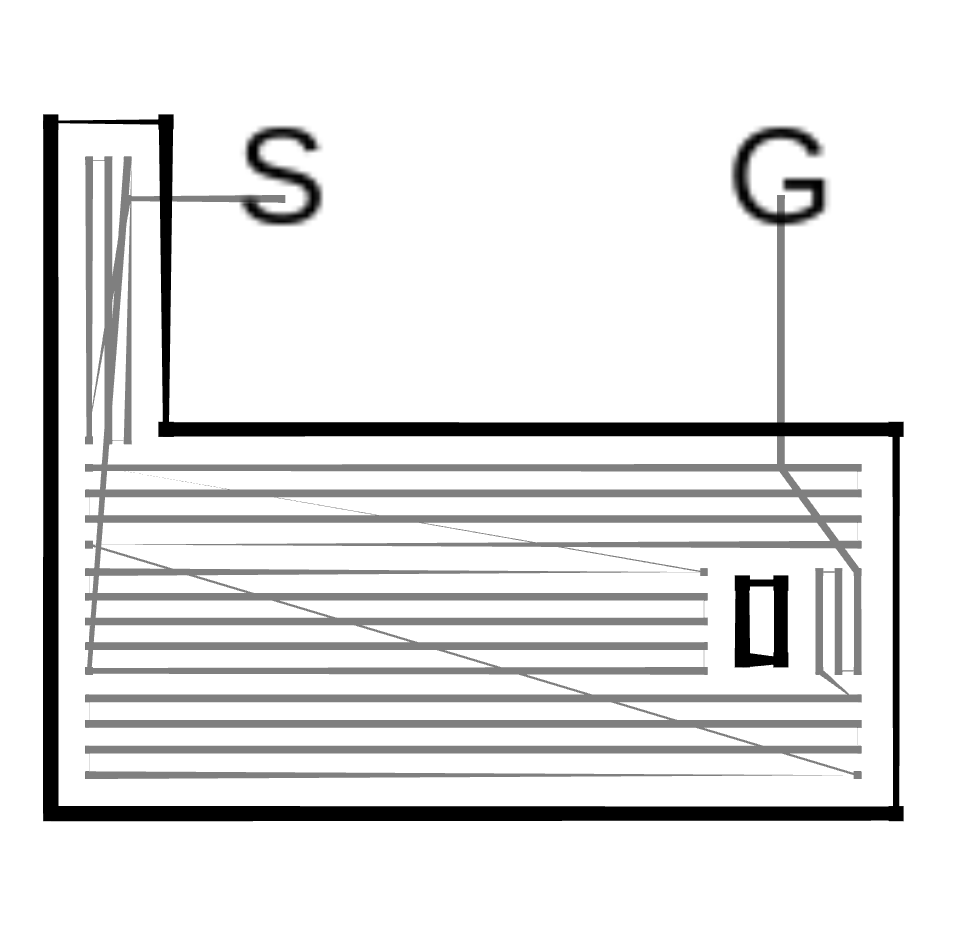} &  
          \includegraphics[height=0.9in]{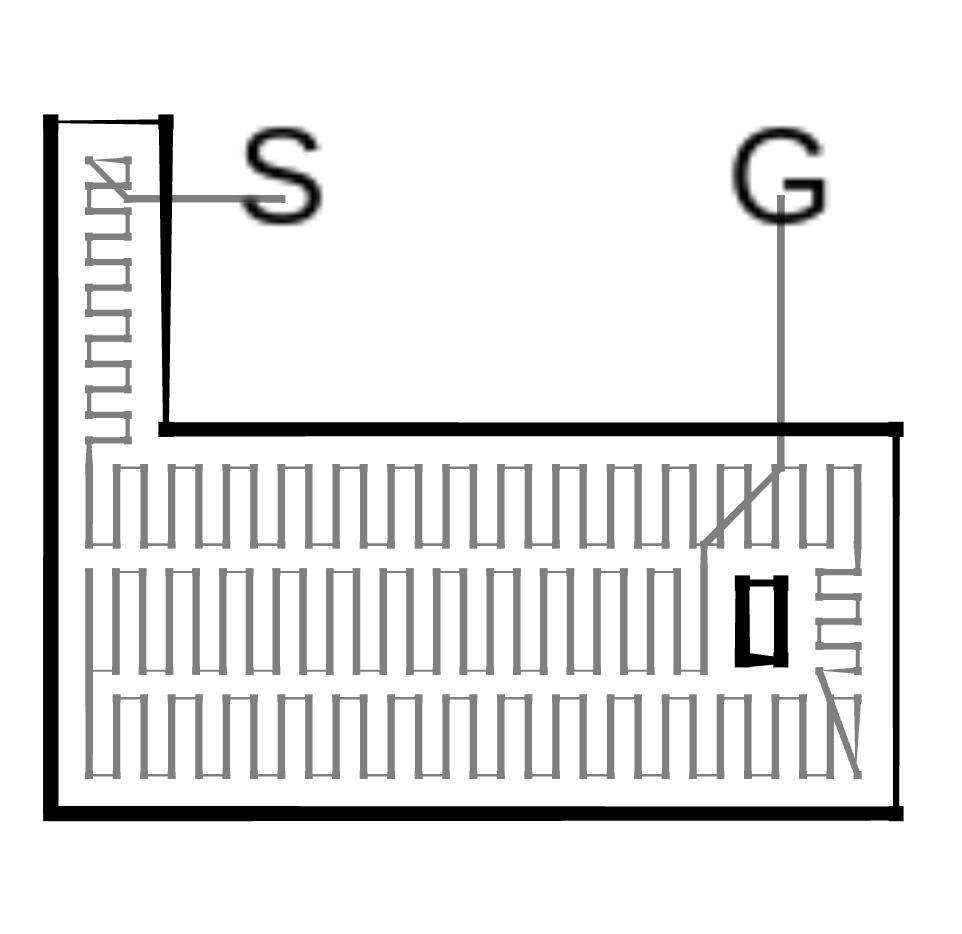} \\[6pt]
          (a) Time & (b) Waypoints & (c) Distance
    \end{tabular}
    \caption{Qualitative comparison of different optimization criteria.
    Minimizing time results in elongated trajectories with short transition segments.
    Minimizing the number of segments also leads to long trajectories, but does not necessarily lead to good transitions.
    Minimizing only the Euclidean distance can lead to undesired sweeps since turns are not penalized.}
    \label{fig:cost}
\end{figure}

Since \acp{MAV} can hover and turn on the spot, we can plan rest-to-rest segments between waypoints which obey the straight-line assumption of our coverage planner.
The trajectories are modeled with velocity ramp profiles with instantaneous acceleration and deceleration and a maximum velocity.
The cost of a trajectory is the sum of all segment times.
The segment time $t$ between two waypoints is a function of the distance $d$, the maximum velocity $v_{max}$, and the maximum acceleration $a_{max}$,
\begin{align}
    t   &= 
    \begin{cases} 
        \sqrt{\frac{4d}{a_{max}}}, & \text{for } d < 2 d_a \\
        2 t_a + \frac{d - 2 d_a}{v_{max}}, & \text{for } d \geq 2 d_a 
    \end{cases},\quad\text{where }
    t_a = \frac{v_{max}}{a_{max}},\quad d_a = \frac{1}{2} v_{max} t_a.
\end{align}
$t_a$ is the time to accelerate to maximum velocity and $d_a$ is the distance travelled while accelerating.
\autoref{fig:cost} shows that by tuning $v_{max}$ and $a_{max}$ with respect to our platform and sensor constraints the optimization finds a good compromise between minimizing distance and minimizing the number of turns.

\autoref{lst:adjacency_graph} summarizes the process of setting up the \ac{E-GTSP} problem.
First we decompose the polygon into monotone cells.
For each cell we compute the sweep permutations and make each sweep pattern a node in the graph where the neighborhood is defined by the cell id.
Finally, we create the edges between all sweeps patterns of different cells using the precomputed reduced visibility graph.
Once the cost matrix is fully defined, we solve the \ac{E-GTSP} using \ac{GK}\footnote{\url{https://csee.essex.ac.uk/staff/dkarap/?page=publications&key=Gutin2009a}} as an off-the-shelf open source solver \cite{gutin2010memetic}.
\begin{algorithm}
\caption{Generating the \Ac{E-GTSP} adjacency graph} \label{lst:adjacency_graph}
\begin{algorithmic}[1]
    \Require $\mathrm{pwh}, \mathrm{sweep\_distance}, \mathrm{wall\_distance}, \mathrm{cost\_func}, \mathrm{decomposition\_type}$
    \Ensure $\mathrm{adj\_graph}$
    \LineComment{Decompose polygon into monotone cells.}
    \State $\mathrm{pwh} = \mathrm{pwh}.\textproc{offsetPolygon}(\mathrm{wall\_distance})$
    \State $\mathrm{cells} = \mathrm{pwh}.\textproc{computeDecomposition}(\mathrm{decomposition\_type})$
    \LineComment{Compute all sweep patterns per cell and create \ac{E-GTSP} nodes.}
    \State $\mathrm{vis\_graph} = \mathrm{pwh}.\textproc{computeVisiblityGraph}()$
    \ForAll {$\mathrm{cell}\in\mathrm{cells}$}
        \State $\mathrm{sweep\_patterns} = \mathrm{cell}.\textproc{computeSweepPatterns}(\mathrm{sweep\_distance}, \mathrm{vis\_graph})$ 
        \ForAll {$\mathrm{sweep\_pattern}\in\mathrm{sweep\_patterns}$}
            \State $\mathrm{n}.\mathrm{sweep\_pattern} = \mathrm{sweep\_pattern}$
            \State $\mathrm{n}.\mathrm{cell\_id} = \mathrm{cell}.\mathrm{id}$
            \State $\mathrm{n}.\mathrm{start\_vis} = \mathrm{pwh}.\textproc{computeVisibility}(       \mathrm{sweep\_pattern}.\mathrm{start})$
            \State $\mathrm{n}.\mathrm{goal\_vis} = \mathrm{pwh}.\textproc{computeVisibility}(       \mathrm{sweep\_pattern}.\mathrm{goal})$
            \State $\mathrm{adj\_graph}.\textproc{addNode}(\mathrm{n})$
        \EndFor
    \EndFor
    \LineComment{Densely connect all nodes via Euclidean shortest paths.}
    \State $\mathrm{adj\_graph}.\textproc{prune}(\mathrm{vis\_graph})$
    \ForAll {$\mathrm{from}\in\mathrm{adj\_graph}.\mathrm{nodes}$} 
        \ForAll {$\mathrm{to}\in\mathrm{adj\_graph}.\mathrm{nodes}\land\mathrm{from}.\mathrm{cell\_id}\neq\mathrm{to}.\mathrm{cell\_id}$}
            \State $\mathrm{e}.\mathrm{path} = \mathrm{vis\_graph}.\textproc{solve}(\mathrm{m}.\mathrm{sweep}.\mathrm{goal}, \mathrm{m}.\mathrm{goal\_vis}, \mathrm{n}.\mathrm{sweep}.\mathrm{goal}, \mathrm{n}.\mathrm{start\_vis})$
            \State $\mathrm{e}.\mathrm{cost} = \mathrm{cost\_func}.\textproc{compute}(\mathrm{n}.\mathrm{sweep\_pattern}) + \mathrm{cost\_func}.\textproc{compute}(\mathrm{e}.\mathrm{path})$
            \State $\mathrm{adj\_graph}.\textproc{addEdge}(\mathrm{e})$
        \EndFor
    \EndFor
\end{algorithmic}
\end{algorithm}

\subsection{Pruning}
While our problem size (tens of clusters, hundreds of nodes, hundreds of thousands of edges) is no problem for the \ac{GK},
generating the edges is the bottleneck of the algorithm because every sweep pattern needs to be connected to almost any other sweep pattern through a Euclidean shortest path.
The total number of arcs grows quadratically with the number of nodes.

Fortunately, the optimization problem is modular, i.e., any sweep pattern combination that visits every cell will achieve full coverage and the path cost is the only optimization criterion.
We can safely prune a node $n_i \in N_k$ if it is cheaper to first traverse from $n_i$'s start point to the start point of $n_j \in N_{k}$, perform coverage $n_j$ and return to $n_i$'s goal point.

\section{Results}
\label{sec:results}
The algorithm has been implemented in C++ using \ac{CGAL} \cite{cgal:eb-18b}.
\ac{CGAL} provides efficient, reliable, and exact geometric algorithms which we extended to generate the sweep lines and the \ac{BCD}.
In order to find an exact solution we implement \cite{rice2012exact} to convert the \ac{E-GTSP} into a directed graph for an exhaustive exact solution using Dijkstra.

\subsection{Simulation Benchmark}
We setup a simulation benchmark to evaluate the performance of our algorithm (\textit{our}) against the classical sweep planner with one sweep direction (\textit{one\_dir}) \cite{choset1998coverage}, to compare \ac{BCD} and \ac{TCD}, and to demonstrate the superiority of a designated \ac{E-GTSP} solver over \textit{exact} brute-force search.
Our test instances, e.g. \autoref{fig:pipeline}, are automatically generated from the EPFL aerial rooftop dataset \cite{sun2014free} to provide polygon maps with realistic obstacles and dimensions.
Our synthetic dataset consists of $320$ rectangular worlds with an area of \SI{1}{\hectare} and $0$ to $15$ rooftop obstacles.
The benchmark was executed on an Intel\textregistered Core\texttrademark i7-7820HQ CPU @ \SI{2.90}{\giga\Hz}.
\begin{figure}
    \centering
    \begin{tabular}{c}
          \includegraphics[width=0.99\textwidth]{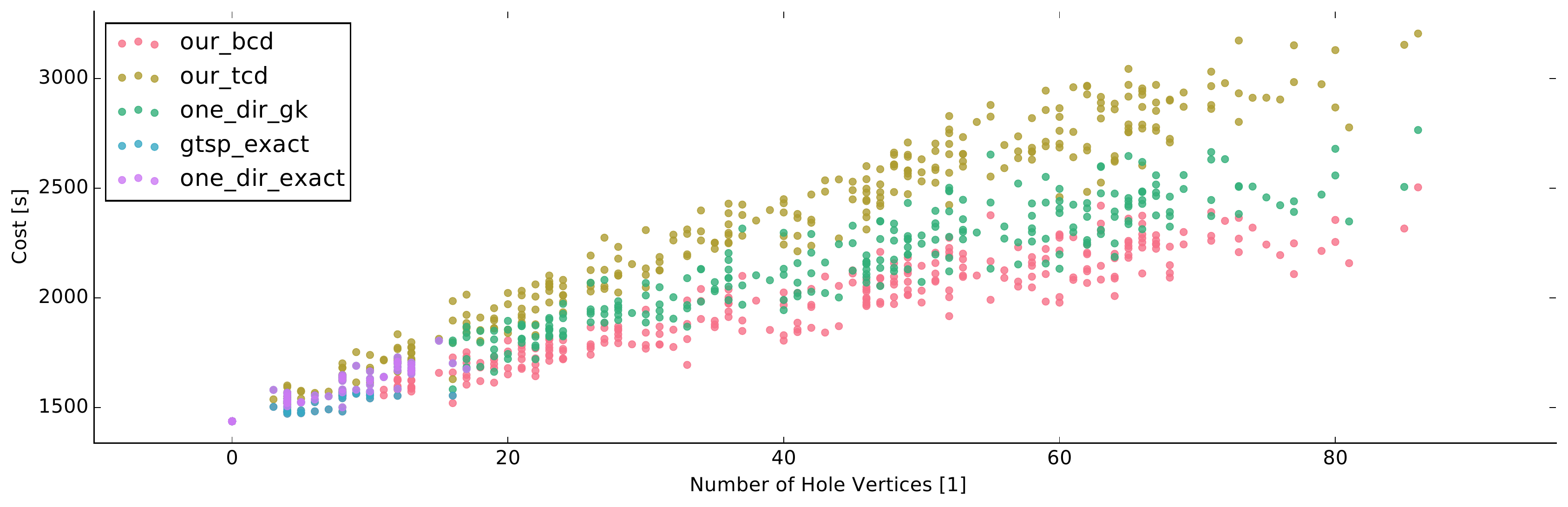} \\[6pt]
          (a) The absolute path cost.\\[6pt]
          \includegraphics[width=0.99\textwidth]{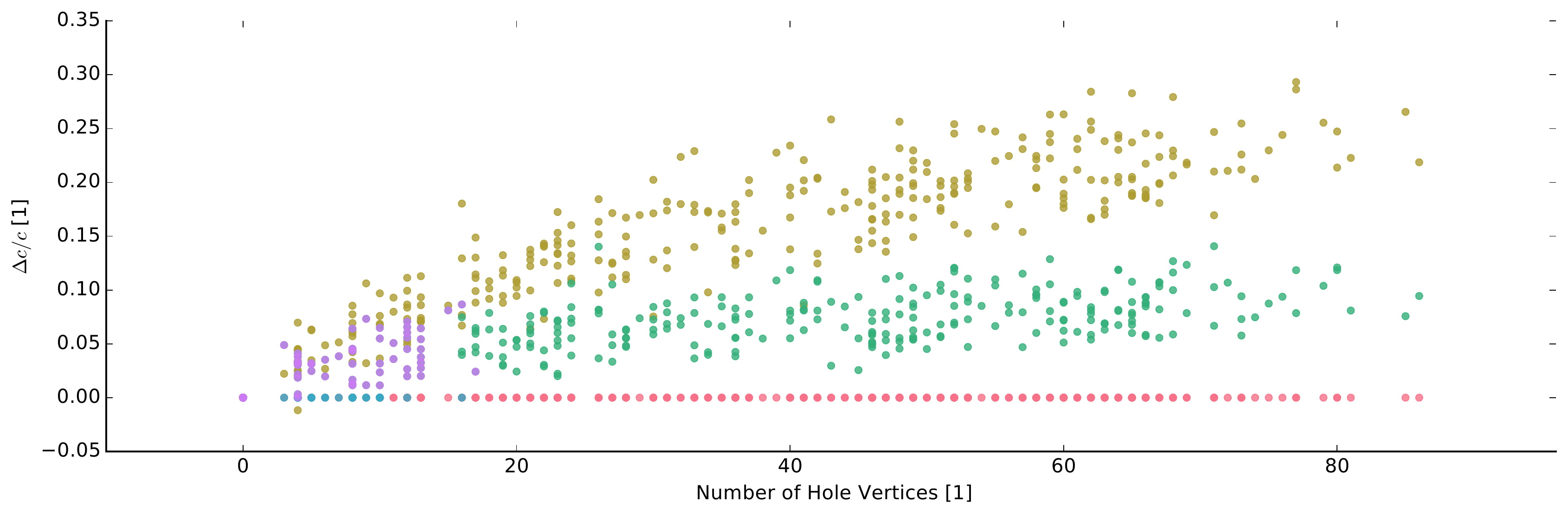} \\[6pt]
          (b) The relative improvement of our planner.\\[6pt]
          \includegraphics[width=0.99\textwidth]{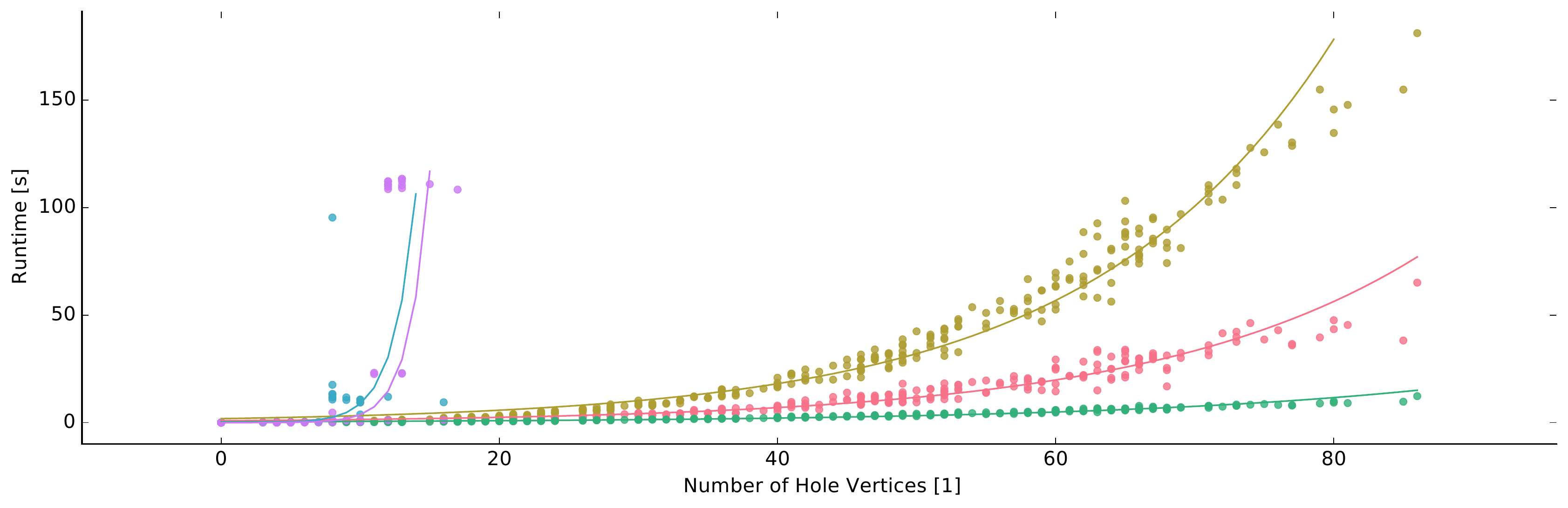} \\[6pt]
          (c) The computation times.
    \end{tabular}
    \caption{Benchmark results for increasing map complexity.
            Our \ac{E-GTSP} with \ac{BCD} generally has the lowest path cost and reasonable computation times even for complex maps.
            Exact solutions only work for maps with few holes.
            The \ac{TCD} is generally worse than \ac{BCD}.}
    \label{fig:benchmark}
\end{figure}

To relate our solution to the complexity of the maps, we plot coverage path cost and algorithm runtime against the number of hole vertices, as these are the events of the decomposition scan line algorithms.
\autoref{fig:benchmark} (a) shows the trajectory cost of the different planner configurations.
Because our planner takes advantage of different sweep directions and start points, it gives better results than the classic solution with only one fixed sweep pattern per cell.
Furthermore, \ac{BCD} leads to shorter trajectories than \ac{TCD} because it generates fewer cells and thus fewer redundant sweeps at the adjacent edges and transition segments between the cells.
On a side note, whenever the exact solution is available it coincides with the approximate solution from \ac{GK}.

The relative differences in path cost between our planner and the other configurations is shown in \autoref{fig:benchmark} (b). We observe up to \SI{14}{\percent} improvement over the optimal fixed direction and \SI{29}{\percent} improvement of \ac{BCD} over \ac{TCD}. 
\autoref{fig:benchmark} (c) shows the computation time of the planners.
Exact solutions fail to solve within \SI{200}{\second} for any of our scenarios with more than twenty hole vertices.
Generating the product graph from the adjacency graph $G$ and Boolean lattice of possible cluster sequences grows exponentially in the number of clusters \cite{rice2012exact}.
The reduced problem \textit{one\_dir\_gk} achieves the best computation time. 
At most it takes \SI{12}{\second} for the most complex map.
The full \ac{E-GTSP} with \ac{BCD} solves it within \SI{65}{\second} which is still reasonable for employing the planner in field experiments.
\autoref{tab:comp} reveals that generating the dense \ac{E-GTSP} graph is the greatest computational burden.
\begin{table*}
\centering
\begin{tabular}{@{}lrrrrrrrrrrrr@{}}
  \toprule
   & \multicolumn{3}{c}{Graph elements} & & \multicolumn{6}{c}{Computation time (s)} & & Path cost (s) \\ 
   \cmidrule{2-4} \cmidrule{6-11} 
     & Cells & Nodes & Edges & & Cells & Sweeps & Nodes & Pruning & Edges & Solve & \\ \midrule
   our\_bcd  & \textbf{\SI{52}{}} & \SI{254}{} & \SI{63056}{} & & \SI{5.68028}{} & \SI{1.100210}{} & \SI{2.984410}{} & \SI{2.392840}{} & \SI{22.2329}{} & \SI{1.195960}{} & & \textbf{\SI{2391.21}{}}  \\
   our\_tcd  & \SI{87}{} & \SI{512}{} & \SI{258704}{} & & \SI{6.92008}{} & \SI{1.107450}{} & \SI{4.940300}{} & \SI{3.679660}{} & \SI{84.6797}{} & \SI{6.555060}{} & & \SI{2879.15}{}  \\
   one\_dir\_gk  & \textbf{\SI{52}{}} & \textbf{\SI{52}{}} & \textbf{\SI{2652}{}} & & \SI{5.53508}{} & \SI{0.089057}{} & \SI{0.246038}{} & \SI{0.026695}{} & \textbf{\SI{0.8067}{}} & \SI{0.513497}{} & & \SI{2664.98}{}  \\
  \bottomrule
\end{tabular}
\caption{Detailed computation times for the instance in \autoref{fig:pipeline} with 15 holes ($52$ hole vertices). Creating all edges is the greatest computational effort. The reduced problem thus has the smallest computation time. Our planner results in the lowest path cost.}
\label{tab:comp}
\end{table*}

\subsection{Experiment}
\begin{figure}
    \centering
    \begin{tabular}{cc}
          \includegraphics[height=3.5cm]{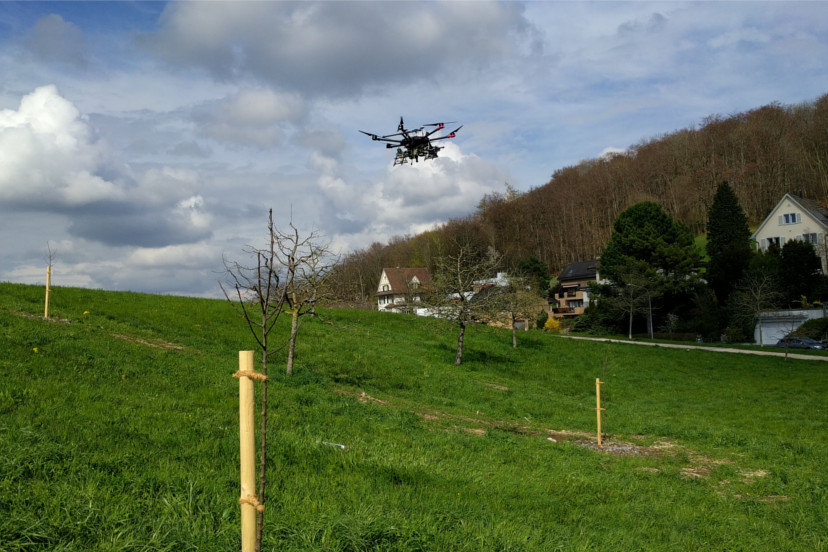} &
          \includegraphics[height=3.5cm]{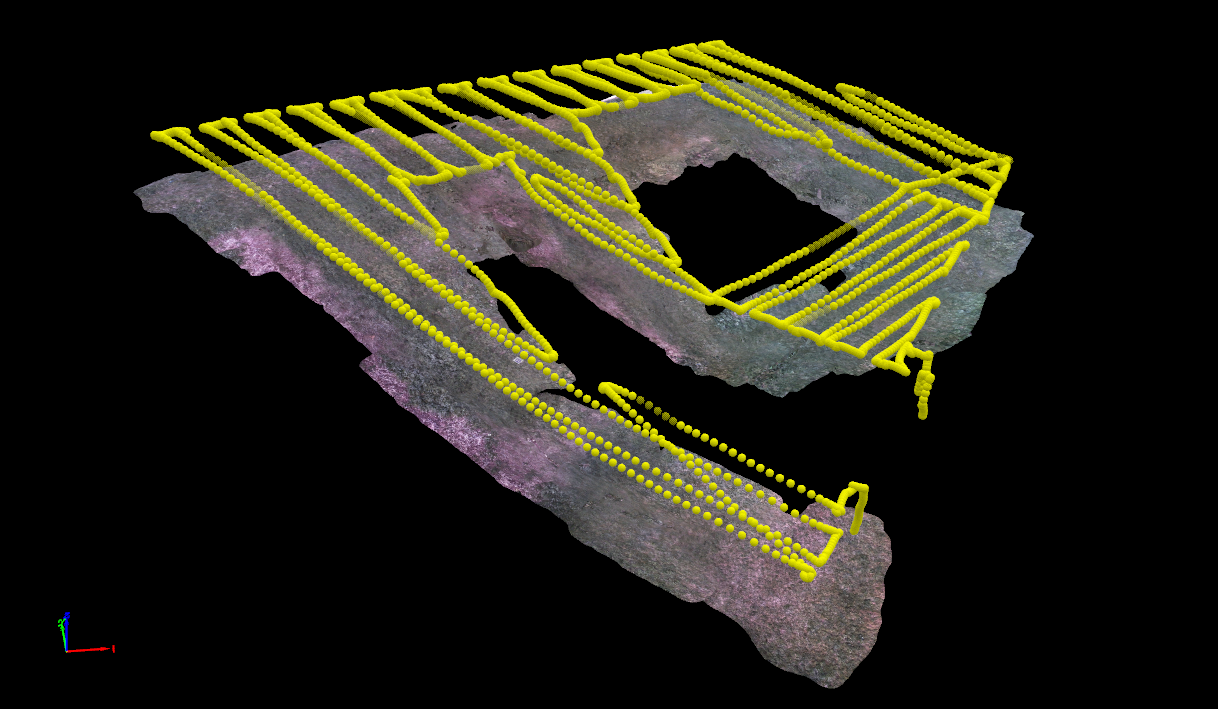} \\[6pt]
          (a) Terrain following using altimeter data.  &
          (b) A Pix4D \ac{DTM} reconstruction. 
    \end{tabular}
    \caption{The platform is capable of covering sloped terrain using the proposed planner.}
    \label{fig:dtm}
\end{figure}

We validate our planner in a real flight on a DJI M600 Pro.
Our drone covers a non-convex open area with sparse trees, depicted in \autoref{fig:eyecatcher}, in a low altitude mapping scenario. 
The flight corridor is selected in a georeferenced map, an optimal sweep path is calculated with our planner and executed under GNSS control.
The drone follows a velocity ramp profile as described in \autoref{sec:gtsp} and additionally turns at every waypoint in the flight direction.
The slope of the terrain is not considered during planning.
To regulate the \ac{AGL}, we fuse Lidar and Radar altimeter data into a consistent altitude estimate.
\autoref{tab:exp} shows the general experiment setup.
The sweep distance is chosen based on image overlap.
The velocity and acceleration are chosen to meet controller constraints and avoid motion blur.
To validate coverage with a nadir configuration sensor we record QXGA top-down imagery and generate the \ac{DTM} shown in \autoref{fig:dtm} using the Pix4D mapping tool.
The \ac{DTM} shows good coverage of the designated area but imperfect reconstruction due to instantaneous acceleration movements.
Collision-free trajectory smoothing may improve the turning maneuvers both in speed and smoothness and consequentially increase overall performance of the coverage planner.
\begin{table*}
\centering
\begin{tabular}{>{\centering}m{1.75cm} >{\centering}m{1.75cm} >{\centering}m{1.75cm} >{\centering}m{1.75cm} >{\centering}m{1.75cm} >{\centering}m{1.75cm}}
  \toprule
    Area & Flight Time & \ac{AGL} & Sweep Offset & $v_{\mathrm{max}}$ & $a_{\mathrm{max}}$ \tabularnewline\midrule
   \SI{1950}{\metre\squared} & $\SI{1050}{\second} $ & \SI{4}{\metre} & \SI{1.5}{\metre} & \SI{3.0}{\meter\per\second} & \SI{0.5}{\meter\per\second\squared} \tabularnewline 
  \bottomrule
\end{tabular}
\caption{Flight experiments parameters.}
\label{tab:exp}
\end{table*}

\section{Conclusion}
\label{sec:conclusions}
In this work we presented a boustrophedon coverage path planner based on an \ac{E-GTSP} formulation.
We showed in comprehensive benchmarks on realistic synthetic polygon maps that our planner reliably solves complex coverage tasks in reasonable computation time, making it suitable for field deployment.
Furthermore, we showed that our planner outperforms the classic boustrophedon coverage planner in terms of path cost.
We validated in a field experiment the usability of our coverage algorithm on a real \ac{MAV} and show that we can cover a \SI{1950}{\metre\squared} area with obstacles at low altitude.
Future work includes optimizing coverage for a side looking \ac{SAR} configuration, collision-free trajectory smoothing to improve turning times, and integration into an airborne mine detection system.

\section*{Acknowledgment}
\label{sec:acknowledgment}
This work is part of the FindMine project and was supported by the Urs Endress Foundation.
The authors would like to thank their student Lucia Liu for her initial work on the \ac{BCD}, Florian Braun and Michael Riner-Kuhn for their hardware support, and the reviewers for their constructive advice.


\bibliographystyle{spmpsci}
\bibliography{bibliography.bib}
\end{document}